# LLM Enhancement with Domain Expert Mental Model to Reduce LLM Hallucination with Causal Prompt Engineering

Boris Kovalerchuk
Central Washington University
USA
borisk@cwu.edu

Brent D. Fegley
Aptima Inc.
USA
bfegley@aptima.com

## Abstract

When consequential decisions depend on knowledge that exists nowhere in writing, LLMs hallucinate not from retrieval failure but from model absence. RAG and knowledge-graph methods share a structural ceiling: they cannot supply what has never been recorded. This paper reconceptualizes that failure as a model discovery problem. The proposed causal prompt engineering framework discovers and encodes a domain expert's decision logic as an Expert Mental Model (EMM), providing it as principled input to LLM reasoning—supplying what no retrieval method can.

The framework is grounded in the theory of monotone Boolean and *k*-valued functions. A key technical contribution is the formalization of three LLM-assisted prerequisite processes previously undeveloped for novel domains: factor formulation, factor monotonization, and monotonicity-preserving hierarchical structuring. These enable a four-step EMM construction algorithm that reduces elicitation from millions of combinations to a tractable, formally bounded set manageable by humans.

The framework also enables, for the first time, formal measurement of LLM hallucination when expert knowledge is entirely tacit. Analysis shows conditional error rates on true positives can approach 100% for plausible LLM-expert mismatches—invisible to prior evaluation methods. The approach is validated across grant proposal evaluation, cybersecurity design, and clinical diagnosis.

# 1 Introduction

## Motivation

Difficult decisions arise constantly in science, engineering, medicine, finance, and countless other fields. Increasingly, practitioners are turning to large language models (LLMs) for support with these decisions. LLMs are impressively capable: they synthesize information, generate options, and explain complex topics in plain language. However, they have a critical weakness. An LLM can only work with information it encountered during training. When a question requires knowledge absent from that training data, the model does not simply admit ignorance—it generates a confident but potentially false answer, a phenomenon known as hallucination.

One established remedy is Retrieval-Augmented Generation (RAG), which supplements an LLM's knowledge by searching external documents before generating a response. Related techniques—knowledge-graph-augmented RAG (KG-RAG) and in-context learning—further enrich the information available at query time. These methods, however, share a fundamental limitation: they can only draw on knowledge that has been recorded somewhere. Many of the most consequential decisions involve knowledge that exists nowhere in writing. A research team evaluating a funding opportunity, an organization designing a novel cybersecurity framework, a clinician synthesizing an unusual pattern of diagnostic indicators—in each case, critical knowledge resides in the minds of experienced practitioners whose judgment no database fully captures. Even a personal question like “What paint color should I buy for my living room?” exposes this gap: without knowing the asker’s tastes, existing furniture, and lighting conditions, the LLM either fabricates a specific recommendation or retreats to a generic non-answer. The same ceiling constrains autoformalization—now extended from translating mathematical statements into machine-verifiable form to rendering arbitrary natural language as verifiable logic for reasoning, planning, and knowledge representation [Mensfelt et al., 2026]—because it can only formalize what language has already captured, whereas the decision model this paper targets was never put into words.

Prompt engineering techniques such as Direct Preference Optimization (DPO) attempt to align LLM behavior more closely with human preferences, but they face an analogous ceiling: a prompt can only convey what the person writing it already knows to include, and fine-tuning on general preference data cannot substitute for the judgment of a specific expert in a specific domain. When the required reasoning belongs to an expert whose mental model is complex, partly intuitive, and difficult to fully articulate, subtle but critical dimensions of judgment are lost in translation.

This paper’s central conceptual move is to reframe LLM hallucination—in the class of decisions just described—as a failure of model discovery rather than a failure of retrieval or alignment. When an LLM hallucinates about a particular team’s strategic priorities, a specific organization’s security posture, or an individual clinician’s diagnostic reasoning, the deficiency is not that the wrong document was retrieved or that the model is poorly calibrated to generic preferences. The deficiency is that the required decision model—the structured encoding of a specific expert’s judgment—exists nowhere the LLM can access. The solution is accordingly not to improve retrieval or refine alignment but to discover and encode that model directly, then supply it as structured, principled input to the reasoning process.

To anchor the discussion, we use a consistent running example throughout: a research team evaluating whether to respond to a funding agency’s call for proposals. This scenario exemplifies a broad class of real decisions that are consequential, time-pressured, involve multiple stakeholders, and lack the historical data that would allow an LLM to reason about them reliably on its own.

## Approach Overview

We propose a framework for constructing an *Expert Mental Model* (EMM)—a structured, computational representation of how a domain expert makes decisions in a given context. Rather than asking an expert to answer an exhaustive battery of questions upfront, the framework uses an optimized dialogue between human and machine to elicit decision logic as efficiently as possible. The resulting model can subsequently be queried on new scenarios without further expert involvement, effectively encoding tacit knowledge in a portable and reusable form.

The framework draws on prior theoretical and applied work [Kovalerchuk et al., 1996, 2001, 2012, 2024; Huber, 2025] and integrates RAG to reduce LLM hallucinations throughout the process. Its mathematical foundation lies in the theory of monotone Boolean and $k$-valued functions, which provides principled techniques for minimizing the number of questions an expert must answer without sacrificing the completeness or fidelity of the resulting model. The framework proceeds in four steps, summarized in Algorithm 1.

**ALGORITHM 1: EXPERT MENTAL MODEL CONSTRUCTION**

**Input:** A description of the decision to be made and the set of valid response values (e.g., yes/no, or a ranked scale such as unacceptable / unsure / acceptable).

**Output:** An Expert Mental Model (EMM) encoding the domain expert's decision logic.

1 **Identify factors.** With LLM assistance, generate a comprehensive list of the factors (attributes) most relevant to the decision. Unlike classical machine learning, where this step is performed entirely by hand, the LLM acts as a collaborative partner to surface factors the expert might not otherwise think to articulate.

2 **Build a hierarchy.** Organize the identified factors into a hierarchy, grouping related factors under broader summary questions. The hierarchy should be constructed such that factors build on one another based on the monotonicity principle—ensuring that higher-level factors logically depend on or extend lower-level ones. LLMs generate candidate hierarchies; domain experts review and refine them. This step is essential when many factors are involved—twenty binary factors produce over one million possible combinations, far too many to evaluate exhaustively.

3 **Specify the generalized model.** Present combinations of the high-level summary factors to the domain expert for classification (e.g., acceptable or unacceptable). The theory of monotone Boolean and $k$-valued functions determines which combinations need to be presented for classification, substantially reducing the total number of required elicitations. Optionally, an LLM can propose a general aggregation method grounded in domain knowledge, which the expert then reviews and approves or not.

4 **Construct the model from the specification.** Using the generalized model from Step 3 as a scaffold, extend the model by repeating the classification process for the more detailed factors, with the same minimization principle in place.

Formally, the core problem is as follows. Given a decision domain, a set of factors each taking values in an ordered discrete set, and a domain expert whose judgment function maps any scenario—a combination of factor values—to a classification label drawn from an ordered scale, the objective is to construct an accurate approximation of that function through an optimized elicitation dialogue that minimizes the number of scenarios presented to the expert while preserving the completeness and fidelity of the resulting model. This problem presents three interconnected challenges: the *elicitation challenge*, in which expert knowledge is partly intuitive and not easily articulated; the *combinatorial challenge*, in which even modest numbers of binary factors produce millions of possible input combinations; and the *aggregation challenge*, in which combining factor values into a final decision cannot in general be reduced to simple arithmetic rules such as majority voting or weighted sums.

A critical but previously unaddressed aspect of deploying this theory in novel decision domains is that three prerequisite properties must be **actively engineered** rather than assumed. First, the relevant decision attributes must be identified and articulated through a principled process—what this paper calls *factor formulation*—because no predefined list of factors exists for novel or under-documented decisions. Second, those factors must be *monotonized*: constructed or reframed so that improving any factor cannot plausibly worsen the overall decision outcome. Third, the resulting factors must be organized into a hierarchy that preserves monotonicity across levels—*monotonicity-preserving hierarchical structuring*—so that the efficiency guarantees of the theory propagate from high-level summary questions down through the detailed sub-questions beneath them. Prior applications of this theory operated in settings where these properties were largely presupposed. This paper develops **LLM-assisted methods** for satisfying each requirement systematically, treating all three as explicit, novel methodological contributions that make the mathematical theory **operationally deployable** across the broad class of tacit-knowledge decision problems this work targets.

The output is a model M(E, S) that maps any scenario S based on selected features to a classification label drawn from the expert's E's chosen scale. For binary decisions this label is simply Yes (Y) or No (N); for finer-grained decisions a *k*-point scale—such as {unacceptable, unsure, acceptable}—may be used, with the binary case as a special instance. Once constructed, M(E, S) can be applied to any scenario within its scope without consulting the expert again.

The framework operates in two modes depending on the nature of the available knowledge source. In its **primary mode**, it engages at least one domain expert in an iterative, structured elicitation *dialogue.* The **automated mode** is applicable when relevant expert knowledge has been codified in a policy *document* containing logical statements in natural language. Here an LLM can (1) assist in *extracting* an appropriate factor hierarchy directly from that document and (2) construct (autoformalize) the detailed model using its stated logic, without requiring a human expert throughout the process. The cybersecurity case study in Section 4.2, which draws the factor hierarchy and aggregation logic directly from FIPS 199 [National Institute of Standards and Technology, 2004], illustrates this mode. The framework's scope is explicitly limited to decisions for which the monotonicity property is plausible—decisions for which improving conditions on any factor should not worsen the overall outcome—a constraint that holds naturally for most evaluative and screening decisions but requires justification in domains where trade-offs are genuinely non-monotone.

## Minimizing the Expert's Burden: Dialog Optimization

A central practical challenge is the combinatorial growth of possible factor combinations. With three yes/no factors there are $2^3 = 8$ combinations; with ten factors, more than one thousand; with twenty, over one million. Reducing the number of necessary questions is therefore essential.

The framework exploits monotonicity—the intuitive principle that a decision should only improve, or stay the same, as conditions improve. If a given set of conditions is sufficient to justify a positive decision, then any scenario with equally good or better conditions on every factor should also warrant a positive decision. Conversely, if a scenario is insufficient, any scenario equal to it or worse on every factor is likewise insufficient.

To illustrate with our running example, suppose three supporting questions are asked: (1) Have you read and fully understood the call? (2) Do we have enough resources? (3) Is the approach genuinely innovative? If expert E decides that (N, N, Y)—No to questions 1 and 2, Yes to question 3—is sufficient to continue work on the proposal, so that M(E, (N, N, Y)) = Y, then monotonicity immediately implies that (Y, N, Y), (N, Y, Y), and (Y, Y, Y) must also be acceptable, since each improves on (N, N, Y) in at least one dimension. Those three combinations need never be asked. The savings compound rapidly as the number of factors grows, and Section 3.1 shows how monotonicity, aggregation logic, and question hierarchies can be combined for even greater efficiency.

## Structuring Questions Hierarchically

Even with monotonicity-based optimization, the number of relevant factors can remain large. A hierarchy of questions addresses this by organizing factors into levels of generality, allowing experts to answer broad summary questions first and elaborate only where necessary.

Consider the summary question: “Is our proposed project well aligned with the funding opportunity?” If the expert’s answer is No, the team has a clear signal to halt—no detailed follow-up is needed. If Yes, more specific questions follow: Have you read and understood the call carefully? Does the project address the agency’s stated priorities? Is the approach genuinely innovative? This structure leverages a real cognitive capacity: people can often answer a generalized question without consciously working through every underlying detail. Asked “Do you trust this person?”, most of us answer quickly; specific reasons emerge only when pressed. LLMs are well-suited to constructing such hierarchies—generating candidate summary questions, organizing detailed factors beneath them, and proposing alternative phrasings—while domain experts refine the results to reflect their own understanding of the problem.

This hierarchy motivates a distinction between two related constructs. The Expert Mental Model (EMM) encodes the expert’s overall decision logic across the hierarchy’s levels. The Explanatory Expert Mental Model (2EMM) unpacks that logic in detail, tracing the specific lower-level factors that drove a particular conclusion. When a decision is negative, the 2EMM is essential: it reveals where the gap lies and what would need to change for the outcome to improve. When a team must reach a collective decision, both the EMMs and 2EMMs of individual members are needed—to surface differences in reasoning, identify points of disagreement, and build shared understanding across the group.

## Contributions and Paper Organization

This paper makes the following contributions:

**A novel reconceptualization of LLM hallucination** in tacit-knowledge settings as a model discovery problem, and a causal prompt engineering framework that integrates Expert Mental Models (EMMs) with LLMs to systematically elicit and encode the decision logic that no retrieval or augmentation method can supply. This framing explicitly fills the gap left by RAG and knowledge-graph-augmented approaches, which are structurally bounded by what exists in recorded form and cannot access expert knowledge that has never been written down.

**Three LLM-assisted prerequisite processes**—factor formulation, factor monotonization, and monotonicity-preserving hierarchical structuring—that are necessary to apply the theory of monotone Boolean and *k*-valued functions to novel decision domains but were not previously formalized, together with a four-step EMM construction algorithm that incorporates these processes to discover computationally tractable expert decision models while dramatically reducing the elicitation burden on domain experts. This contribution became possible only with the advent of LLMs.

**Validation across three high-stakes domains** (grant proposal evaluation, cybersecurity framework design, and healthcare) demonstrating the framework’s generalizability across qualitatively distinct knowledge environments—from entirely tacit expert knowledge to doctrine-scaffolded and empirically grounded settings—and showing that the same formal machinery operates appropriately in each.

**Formal, measurable quantification of LLM hallucination** for the class of settings where expert knowledge is absent from all accessible sources—a failure mode not previously quantifiable by any method. By constructing families of Boolean decision functions spanning the realistic range of LLM-generated guesses and deriving explicit error bounds, the analysis yields new empirical findings about LLM weaknesses: conditional error rates on true positives can approach totality for plausible mismatches between LLM statistical priors and the actual expert decision function, even when population-level error rates appear reassuringly small.

The remainder of the paper is organized as follows: Section 2 reviews related work. Section 3 details the methodology and algorithms. Section 4 presents case studies from the three domains. Section 5 presents a discussion and Section 6 concludes with directions for future research.

# Related Work

## Overview

This section surveys five interrelated lines: cognitive analogs and limits of LLMs; classic expert mental-model theory; hybrid knowledge architectures; expert-persona and prompt-engineering methods; and current system limitations. The synthesis in Section 2.7 draws these strands together and connects them to the elicitation approach developed in Section 3. A comparative overview of representative methods appears in the Discussion (Section 5.3, Table 4).

## LLMs as Cognitive Systems

A foundational question is whether LLMs organize knowledge analogously to expert cognition. When Binz and Schulz applied canonical cognitive tasks to GPT-3, they found it matched or exceeded humans on vignette-based decisions but struggled with causal reasoning [Binz & Schulz, 2023]. Sartori and Orrù situate this in relation to associationist theory, positing that self-attention explains LLMs' ability to perform at human levels on tests such as metaphor comprehension and Raven's matrices [Sartori & Orrù, 2023]. A structural analysis of 591 LLMs on twelve tests aligned with fluid reasoning, domain knowledge, and quantitative ability found a positive manifold analogous to the general intelligence factor in humans, with parameter count correlating positively but with diminishing returns [Ilić & Gignac, 2024].

Benchmark methodology is itself contested terrain. LLMs now exceed 90% accuracy on the Measuring Massive Multitask Language Understanding benchmark, limiting meaningful capability differentiation [Center for AI Safety, Scale AI, & HLE Contributors Consortium, 2026]. Humanity's Last Exam (HLE), a multi-modal benchmark of 2,500 questions covering more than 100 subjects, was presented as the final closed-ended academic benchmark; expert humans averaged about 90% [Center for AI Safety, Scale AI, & HLE Contributors Consortium, 2026; Laurent, 2026a]. At release in early 2025, GPT-4o scored 2.7% and o1 scored 8.0%; by early 2026, Gemini 3 Pro Preview led at 37.5%, with a still-substantial gap from expert performance [Humanity's Last Exam, 2026; Laurent, 2026a]. In medical domains, GPT-5 achieved 95.84% accuracy on MedQA and 95.22% averaged across USMLE steps, largely exceeding passing thresholds [Laurent, 2026b]. These trajectories illustrate that benchmark accuracy can approach expert levels without proving the underlying mechanisms are the same.

## Classical Foundations and Expert Knowledge Integration

Classical cognitive science established principles that can potentially inform LLM integration: expert knowledge is structured and causal rather than merely associative; systematic elicitation using multiple representational formats is essential for formalization; and explicit models enable validation and transfer that tacit knowledge cannot. For example, Ford and Sterman's structured elicitation methodology [1998] transforms tacit process knowledge into formal models through sequential phases and multiple representation formats, potentially improving both the artifact and the expert's own understanding. Morgan and colleagues [2002] developed the mental models approach to risk communication, representing expert causal structures as influence diagrams and comparing them to lay models to guide communication interventions—a framework potentially adaptable for grounding LLM outputs in validated domain knowledge.

Recent clinical work demonstrates the practical value of explicitly encoding these structures. In a study by Sarma et al. [2026], direct LLM prompting for psychiatric diagnoses from case vignettes produced most true positives but significant overdiagnosis; integrating expert-derived decision trees reportedly improved F1 primarily by suppressing false positives. More strikingly, Rollwage and colleagues [2026] reportedly introduced a cognitive layer architecture augmenting LLMs with psychotherapeutic reasoning; in a claimed randomized double-blind evaluation, the augmented system outperformed standalone LLMs and human clinicians on cognitive-behavioral therapy (CBT) competencies.

## Cognitive Architectures, Knowledge Graphs, and Hybrid Systems

Wu and colleagues [2025] present Cognitive LLMs, a hybrid architecture reportedly coupling ACT-R and an LLM through a knowledge transfer mechanism modeling decision profiles in manufacturing. Sun [2024] argues more broadly for a dual-process, hybrid neuro-symbolic approach, contending that future systems must be structured in a cognitively motivated way grounded in human mental processes rather than purely emergent from scale.

Knowledge graph (KG) integration offers a complementary path for grounding LLM generation in structured expert knowledge. LLMs can enable generative knowledge modeling, semantic unification across heterogeneous sources, and instruction-driven orchestration of KG construction workflows, as explored in recent works such as Bian [2025]. Integration paradigms include KG-augmented LLMs, LLM-augmented KGs, and synergized frameworks, leveraging LLMs' contextual understanding to improve knowledge representation accuracy and adaptability [Ibrahim et al., 2024]. Augmenting LLMs with psychologically grounded causal reasoning models, as in Basu et al. [2026], can specifically address planning under uncertainty where purely statistical inference fails.

## Expert Role, Persona Prompting, and Expert-in-the-Loop Methodologies

Through generations of refinement, prompt engineering evolved from an informal art into a systematic discipline, enabling partial translation of expert mental models into LLM behavior without fine-tuning. Shanahan et al. argue that role specification triggers domain-knowledge clusters [Shanahan, McDonell, & Reynolds, 2023]. Kong et al. found measurable performance gains when role descriptions were specific [Kong et al., 2024]. However, Salewski et al. caution that vague labels provide little genuine scaffolding—in-context impersonation reflects the statistical properties of expert discourse in training data rather than structural expert reasoning [Salewski et al., 2023].

More systematic approaches deploy curated schemas, hierarchical prompts, and explicit expert decision structures to reduce hallucinations and improve verifiability [Martin-Boyle et al., 2026]. Strobelt and colleagues demonstrated that small prompt additions by domain experts yield substantial performance improvements [Strobelt et al., 2023]. Decomposing expert workflows into chained LLM calls governed by narrowly scoped prompts translates multi-step expert procedures into maintainable pipelines [Wu et al., 2022], which is actively used in emerging agentic LLM systems. Building prompts requires a clear understanding of what the LLM knows [Reynolds & McDonell, 2021]. Automatic prompt-optimization methods—including Automatic Prompt Engineer [Zhou et al., 2023] and DSPy [Khattab et al., 2023]—still depend on expert evaluation criteria and task decompositions, confirming that expert mental models are pushed upstream rather than eliminated. An important meta-level evolution involves developing expert schemas of LLM behavior itself: Martin-Boyle et al. elicited a schema of LLM error patterns— hallucinations, citation errors, concept conflation—from domain scientists in scholarly question-answering tasks, producing a meta-level mental model of LLM failure modes for systematic evaluation [Martin-Boyle et al., 2026].

## Theory of Mind, Limitations, and Systemic Risks

Expert clinical, legal, and educational cognition requires not only declarative knowledge but theory of mind (ToM). Testing GPT and LLaMA2 families on ToM tasks shows progress on false-belief tasks but mixed performance across benchmarks, as recent studies indicate. Crucially, LLMs may pass ToM benchmarks due to how the models were represented in their training data without spontaneously computing mentalistic inferences under uncertainty—a gap between what models know and what they do in practice. [Bonnefon & Rahwan, 2020; Kosinski, 2024; Jones, Trott, & Bergen, 2024; Strachan et al., 2024]

Hallucination—the production of factually incorrect outputs—is a prominent operational limitation of LLMs, with expert domains disproportionately vulnerable; knowledge grounding methods (such as KG integration) can help mitigate this [Lavrinovics et al., 2025]. In mental health contexts, where LLMs work by recognizing patterns rather than truly understanding content, spreading false information is a documented risk [Guo et al., 2024]. A comparison of LLMs against expert suicidologists found upward bias in judging responses as appropriate and variable performance across models, underscoring that narrow benchmark success does not guarantee reliable clinical utility [Rousmaniere, Goldberg, & Torous, 2026].

A systemic risk identified in recent theoretical work is cognitive homogenization: as LLMs become embedded in professional workflows, they risk standardizing reasoning by reflecting and reinforcing dominant styles while marginalizing alternative reasoning strategies, threatening the cognitive diversity that drives collective intelligence and adaptability in expert communities [Sourati, Ziabari, & Dehghani, 2026].

## Synthesis and Future Directions

Literature converges on several conclusions that directly motivate the approach developed in this paper. LLMs show increasingly sophisticated capabilities in language processing, reasoning, and problem-solving and have become a focal point for examining machine cognition in relation to human expertise [Niu et al., 2024], yet they operate through mechanisms categorically different from genuine expertise—lacking the grounded, causal, metacognitive structures that define domain mastery. The productive response is hybrid architecture that integrates expert models [Sarma et al., 2026; Rollwage et al., 2026], integrated cognitive architectures [Wu et al., 2025], knowledge graphs [Ibrahim et al., 2024; Bian, 2025], or psychologically grounded causal reasoning models [Basu et al., 2026]. Prompt engineering functions as a bridge between expert mental models and LLM capabilities—expert reasoning is partially activated through persona assignment and hierarchical prompts [Shanahan, McDonell, & Reynolds, 2023; Kong et al., 2024], but it relies on training-data patterns rather than structural reasoning [Salewski et al., 2023], motivating systematic elicitation methodologies drawing on classical expert knowledge formalization traditions [Ford & Sterman, 1998; Morgan et al., 2002]. The broader implications of these conclusions for research directions and practice are taken up in Section 5.7.

# Expert Mining with Monotone Boolean and *k*-valued Functions

## Personal Aggregation Logic

To avoid **combinatorial explosion** when building an expert mental model and its aggregation logic, we propose a method that exploits four key properties: hierarchical grouping of questions into categories, monotonicity, heuristics, and iterative refinement.

*Hierarchical grouping* focuses elicitation on high-level categories first and drills down into specific questions only when necessary. Section 4 explores this concept in detail.

*Monotonicity* is applied as follows. For illustration, assume:

six detailed binary factors $(x_1, x_2, \dots, x_6)$ comprising $2^6 = 64$ questions required to build the mental model, and two high-level categories (themselves binary factors), $w_1$ and $w_2$, each generalizing a disjoint set of three of the six binary factors, yielding four factor combinations: (Y,Y), (N,Y), (Y,N), and (N,N). Here $w_1$ represents $(x_1,x_2,x_3)$ and $w_2$ represents $(x_4,x_5,x_6)$.

If an expert can evaluate $w_1$ and $w_2$ without drilling down to the detailed questions, the number of required elicitations is dramatically reduced. If the expert accepts (N,N) for $(w_1,w_2)$ as acceptable (Y), no further drilling down is necessary. If the expert accepts (Y,N) or (N,Y), then under the monotonicity property (given N < Y), the combination (Y,Y) is also necessarily acceptable and need not be asked.

Consequently, it is only necessary to drill into the affirmative category—into $w_1$ when (Y,N) is selected, or into $w_2$ when (N,Y) is selected—requiring at most $2^3 = 8$ questions over the three sub-factors within that category, while the other category remains fixed at N. Monotonicity further reduces the question count within the drilled category: if the expert accepts the triple (Y,N,N), then (Y,Y,N), (Y,N,Y), and (Y,Y,Y) are automatically acceptable and need not be posed.

*Heuristics* involves generating pilot questions that guide the ordering of subsequent questions to eliminate the greatest number of redundant questions [Huber, Kovalerchuk, 2024].

*Iterative refinement* adjusts the ordering of questions based on answers already obtained. The process begins with a small number of generalized questions and progressively refines the mental model through feedback.

A detailed formalization of the monotonicity approach can be found in [Kovalerchuk et al., 1996; Huber, Kovalerchuk, 2024]. The approach has broad applicability, including medical diagnostics, real estate evaluation, and software enhancement—for example, integration of the monotone function approach into a Microsoft Teams bot with a UI/UX for monotone Boolean and *k*-valued function logic—as well as visualization and visual analytics with monotone Boolean and *k*-valued functions [Kovalerchuk et al., 1996, 2012, 2022; Huber, Kovalerchuk, 2024; Huber, 2025]. Further extension to agent-based applications is also feasible, where multiple agents collaborate on a task and an optimal team configuration must be identified—such as selecting an ensemble of machine learning algorithms to construct a predictive model.

## Worked Example: Monotonicity-based Reduction in the Proposal Submission Domain

To give quantitative grounding for the efficiency gains described above, and to anticipate the full case study in Section 4.1, this subsection traces the elicitation process through a concrete sequence of expert judgments drawn from the proposal submission domain. The factor set is the seven generalized questions constructed with LLM assistance in Section 4.1.4:

$x_1$ — Is the project sufficiently aligned with the funding opportunity and innovative?
$x_2$ — Is the organization sufficiently prepared to undertake this project?
$x_3$ — Is the project feasible within the given constraints?
$x_4$ — Is the project financially viable?
$x_5$ — Do we have sufficient collaborative support for the project?
$x_6$ — Can we meet the compliance requirements?
$x_7$ — Is the project's potential impact sufficiently high and risks manageable?

Each factor takes a binary value (Y or N), so a scenario is an element of $\{N, Y\}^7$. Without any optimization, fully specifying an expert's decision function over these seven factors requires classifying all $2^7 = 128$ distinct combinations. The following four elicitations show how the monotonicity principle reduces this burden substantially.

Two inference rules drive the reduction. *Upward closure:* if scenario S is classified as Y, then every $S' \geq S$, factor by factor, is automatically Y, since improving any factor cannot worsen the decision. *Downward closure:* if S is classified as N, then every $S' \leq S$, factor by factor, is automatically N, since worsening any factor cannot improve the decision. Each direct expert judgment propagates to a potentially large neighborhood of the input space with no further elicitation required.

The initial scenario S for expert elicitation can be selected randomly, systematically by hierarchical level within the factor structure, or strategically based on domain knowledge. For simplicity, we employ random selection. The selection of subsequent factors based on experts' prior responses follows methods described in Kovalerchuk et al. [1996] and Torvik [2002]. We do not detail these mechanics here but note that the approach balances the two inference rules described above.

**Elicitation 1.** The expert is presented with $S_1 = (\mathrm{Y, Y, Y, Y, N, N, N})$: the project is well-aligned and innovative, the team is organizationally ready, technical feasibility has been established, and the budget is within the agency's limits—but no collaborative partners have been secured, compliance planning has not begun, and no impact or risk analysis has been conducted.

*Expert verdict: N.* Without any established collaborative network, compliance plan, or impact strategy, committing two months to the proposal would be premature. The four affirmative factors are necessary but not sufficient.

*Downward closure.* Every $S' \leq S_1$ is also N. This consists of all scenarios in which $x_5, x_6, x_7$ are fixed at N while $x_1$ through $x_4$ take any values—a set of $2^4 = 16$ combinations. One is the elicited scenario; the remaining 15 are classified automatically.

*Running tally: 16 classified (all N); 112 remaining.*

**Elicitation 2.** The expert is presented with $S_2 = (\mathrm{Y, N, Y, N, Y, Y, Y})$: aligned and innovative, technically feasible, collaborative, compliance-ready, and impact-aware—but organizational readiness and financial viability are still being finalized.

*Expert verdict: Y.* Strategic alignment, technical feasibility, established partnerships, compliance capability, and a robust impact and risk plan together provide sufficient basis to commit, even with internal resourcing and budgeting still in progress.

*Upward closure* yields $2^2 = 4$ combinations ($x_1, x_3, x_5, x_6, x_7$ fixed at Y; $x_2$ and $x_4$ free); three are new inferences.

*Running tally: 20 classified (16 N, 4 Y); 108 remaining.*

**Elicitation 3.** The expert is presented with $S_3 = (\mathrm{N, Y, Y, Y, Y, Y, Y})$: every factor is affirmative except project alignment and innovation.

*Expert verdict: N.* Without genuine alignment with the funding call and a novel intellectual contribution, the team should not commit to writing the proposal regardless of how well-resourced it is on every other dimension. In this expert's judgment, alignment is a non-negotiable prerequisite.

*Downward closure.* The downward closure of $S_3$ consists of all $2^6 = 64$ scenarios in which $x_1 = N$ while $x_2$ through $x_7$ take any values. Eight of these overlap with the N block from Elicitation 1, so the net gain is 56 newly classified scenarios.

*Running tally: 76 classified (72 N, 4 Y); 52 remaining.*

**Elicitation 4.** The expert is presented with $S_4 = (\mathrm{Y, Y, N, Y, N, Y, N})$: aligned and innovative, organizationally ready, financially viable, and compliance-ready—but technical feasibility, collaborative partnerships, and a formal impact and risk strategy are still to be developed.

*Expert verdict: Y.* Strategic alignment, institutional readiness, financial backing, and compliance confidence provide a sufficient foundation to commit; technical details, partnership agreements, and impact assessment can be developed during the proposal preparation period.

*Upward closure* yields $2^3 = 8$ combinations ($x_1, x_2, x_4, x_6$ fixed at Y; $x_3, x_5, x_7$ free); one overlaps with the upward closure of $S_2$, so six are new inferences.

*Running tally after all four elicitations: 83 classified (72 N, 11 Y); 45 remaining.* Table 1 records the full counts.

*Table 1. Cumulative reduction through four monotonicity-based elicitations over seven binary factors (2^7=128 total combinations). "Automatically inferred" counts combinations classified by monotone closure, excluding the directly elicited scenario in each row.*

| Elicitation | Scenario evaluated | Expert verdict | Automatically inferred | Cumulative classified | Remaining |
|---|---|---|---|---|---|
| 1 | $(Y, Y, Y, Y, N, N, N)$ | N | 15 | 16 | 112 |
| 2 | $(Y, N, Y, N, Y, Y, Y)$ | Y | 3 | 20 | 108 |
| 3 | $(N, Y, Y, Y, Y, Y, Y)$ | N | 55 | 76 | 52 |
| 4 | $(Y, Y, N, Y, N, Y, N)$ | Y | 6 | 83 | 45 |
| **Total** | — | — | **79** | **83** | **45** |

The ordering of the elicitations is worth noting. Elicitation 3 is by far the most powerful single step due to downward closure; it eliminates 56 scenarios in one judgment by exploiting the necessary-condition status of alignment. In practice, the heuristic of leading with strong scenarios (those close to the all-Y corner) when seeking N verdicts, and with weaker scenarios (those with fewer affirmative factors) when seeking Y verdicts, systematically maximizes the size of the automatically inferred set at each step. The 45 remaining unclassified scenarios are not distributed uniformly across the space; they concentrate near the boundary between the N and Y regions, precisely where each additional elicitation carries the highest discriminating power. Elicitation continues until the boundary has been characterized to a resolution the expert considers adequate, at which point the resulting Expert Mental Model can be applied to new scenarios without further expert involvement.

## Group Aggregation Logic

Combining the mental models of several experts into a consensus-based judgment is a central challenge in group decision-making. Algorithm 2 outlines a step-by-step process tailored to this problem.

**ALGORITHM 2: AGGREGATION OF EXPERT MENTAL MODELS**

***Input:*** *Key questions and scenarios around a decision point, itself a question to be answered.*

***Output:*** *An expert mental model (EMM) based on aggregation of personal mental models.*

**1** *Gather individual preferences for identified key questions and scenarios.*

**2** *Compare personal mental models to identify points of agreement and disagreement.*

**3** *Discuss differences in personal mental models to establish the reasons for any disagreements.*

**4** *Aggregate personal mental models into a group mental model, documenting how decisions were reached.*

As a concrete application of Algorithm 2, consider a scenario (Y,N,Y) with three expert responses: $E_1(Y, N, Y) = Y$, $E_2(Y, N, Y) = N$, and $E_3(Y, N, Y) = Y$. Under **majority voting**, the group judgment is $E_{\{1,2,3\}}(Y, N, Y) = Y$, completing the aggregation.

To ground this illustration, consider the three binary factors as corresponding to the following evaluative questions: $x_1$ = "Is the project well-aligned with the funding opportunity?", $x_2$ = "Do we have sufficient resources to execute the proposed work?", and $x_3$ = "Is the technical approach sufficiently innovative?" Under this labeling, the scenario (Y, N, Y) represents a proposal that is well-aligned and technically innovative but under-resourced. Expert $E_1$'s weighted reasoning about the N component—$x_2$—then carries a specific interpretive meaning: $E_1$ is not simply registering a negative vote on an abstract dimension but is making an argued judgment that the resource gap is either remediable or insufficient to override the proposal's strengths on the other two factors. This is precisely the type of nuanced, dimension-specific reasoning that bare majority voting cannot recover. A majority vote across the three factors would record a 2-to-1 outcome in favor of a positive aggregate assessment without preserving the expert's analysis of why the negative factor was weighted as it was, or whether it was treated as a threshold disqualifier or a compensable deficiency. The named-factor framing makes visible the added value of the reasoning-based aggregation procedure developed in this section.

An alternative **reasoning-based process** applies when the team decides against majority voting and instead asks expert $E_2$—who rejected (Y,N,Y) by answering N—to clarify her reasoning. Suppose $E_2$ assigns a substantially higher weight to the N component of the scenario, making the overall scenario unacceptable regardless of the other factors. In that case, $E_2$'s judgment may prevail if she possesses superior situational or domain awareness. Importantly, this reasoning-based process applies to a specific scenario rather than serving as a universal aggregation rule across all scenarios.

Now suppose the experts conclude from their discussion that the scenario is underspecified, and that adding one question would resolve the ambiguity and produce unanimity. The scenario then expands from three to four elements—e.g., (Y,N,Y,N) in place of (Y,N,Y). At the mathematical level, the experts have reconstructed their individual mental models (functions) with this addition, as proposed in [Huber, Kovalerchuk, 2024]. This example illustrates the advantages a **reasoning-based process** can offer over traditional **majority voting** as an alternative aggregation strategy.

## Visualization Support

A fundamental distinction between the proposed approach and traditional computational aggregation rules is that monotone Boolean and ordinal $k$-valued functions may not admit a compact analytical formula. Visualization therefore provides a natural means of representing this complexity and enabling domain experts to analyze and interact with their mental models. Figure 1 [Kovalerchuk et al., 2012] illustrates such a visualization for all combinations of 11 binary questions (no/yes), depicting two mental models of an expert radiologist applied to mammographic X-ray images: one for breast cancer diagnosis and one for the biopsy decision.

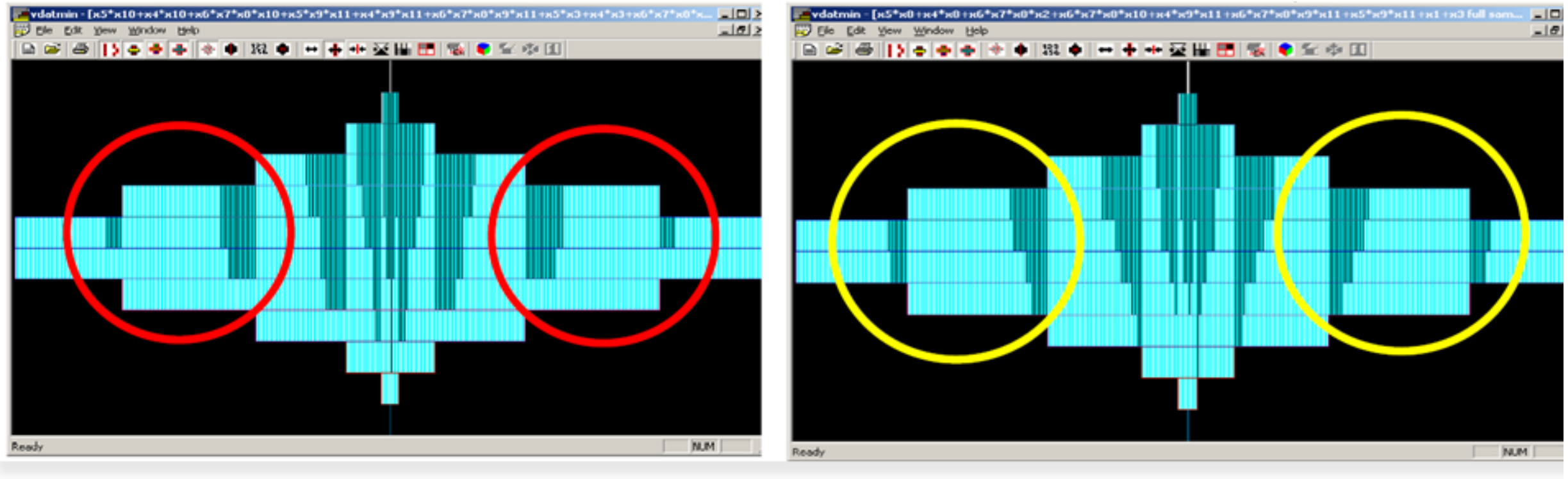

(a) Expert mental model, cancer (b) Expert mental model, biopsy

*Figure 1. Example visualizations of expert mental models [Kovalerchuk et al, 2012].*

All 11-dimensional points representing a "yes" response are shown as black bars, and all points representing a "no" response are shown as white bars. This visualization exploits the concept of Hansel chains for monotone Boolean functions [Kovalerchuk et al., 2012]. The red and yellow circles highlight regions where the two models diverge. For $k$-valued functions with $k > 2$, alternative visualizations are required; these can be constructed by adapting lossless visualization techniques for multidimensional data [Kovalerchuk, 2018; Kovalerchuk et al., 2024].

To illustrate continuity between this methodology and its application in the case studies that follow, consider schematically how the visualization technique would be applied to the proposal submission EMM presented in Section 3.1.1. In that context, the axes of Figure 1 would correspond to the evaluative factors introduced above—alignment, resources, and innovation—with each expert's position plotted according to their weighted factor assessments rather than raw binary votes. The resulting configuration would make legible both the degree of inter-expert agreement on individual factors and the structure of any divergence, distinguishing, for instance, a situation in which experts converge on alignment and innovation but diverge sharply on resources from one in which disagreement is distributed uniformly across factors. Even at this schematic level, the visualization reveals patterns that aggregate scores conceal, and it is this diagnostic capacity that motivates its use. Due to space limitations, this paper does not develop visualization support of this kind further across the case studies in Sections 4 through 6, but the approach is noted here as a potentially useful tool for practitioners and researchers seeking richer interpretive access to the structure of expert disagreement in applied EMM contexts.

## Prompt Engineering Formalization

Recent work in formal prompt engineering has established rigorous frameworks for deploying LLMs as classifiers in well-defined tasks. Cirillo et al. (2025), for instance, propose a formal problem definition that casts cyberbullying detection as a mapping from social media posts to label-plus-explanation pairs under specified prompt conditions, and demonstrate that a carefully engineered, generalizable prompt template enables LLMs to perform competitively with traditional machine learning models—and, in multiclass settings, to outperform them—across a broad comparative evaluation. That said, the paper also finds that fine-tuned transformer-based NLP models (e.g., BERT Large, RoBERTa) outperform LLMs across all tasks, achieving F1 scores on the order of 0.94 on the multiclass benchmark where the best-performing LLM reaches approximately 0.77, a gap the authors attribute to the inherent advantage NLP models derive from task-specific fine-tuning on domain-representative data. The Cirillo et al. formalism nonetheless presupposes an architectural configuration in which the *LLM is the decision-making agent*, prompt engineering is the primary methodological contribution, and LLM outputs can be assessed against a ground truth that exists independently of any individual expert. The present framework operates on a categorically different architecture: the *LLM plays an assistive rather than decisive role*, generating candidate factors for expert review and structuring the elicitation dialogue, while the *expert retains sole authority over every judgment* that enters the model. Furthermore, because each prompt in the elicitation dialogue is conditioned on the expert's prior judgments, the current state of the monotonicity lattice, and the set of scenario combinations that remain unresolved, the prompting interaction is adaptive and state-dependent by design rather than template-uniform—making a fixed prompt schema an inaccurate representation of the process. For these reasons, the framework does not adopt the Cirillo et al. formalism, though that body of work represents an important point of reference for research in which LLM-based classification and prompt design are themselves the objects of formal investigation.

# Case Studies: Proof-of-Concept Demonstrations

Before presenting the individual case studies, it is worth being explicit about the nature and scope of this paper's empirical contributions. The work presented here is primarily a framework and algorithm paper. Its central contributions are methodological and formal: the development of a reasoning-based expert aggregation procedure grounded in monotone Boolean and $k$-valued function families, and the demonstration that this procedure is instantiable across qualitatively different knowledge domains. Many strong papers in AI and decision science do not include large-scale empirical evaluations, and we do not claim to offer one here. What the three case studies do provide is something distinct and appropriate to the paper's ambitions: proof-of-concept demonstrations that the formal machinery developed in Section 3 can be applied to real evaluative problems in domains with meaningfully different knowledge structures. Case study 1 addresses an evaluated decision problem in proposal review; case study 2 engages a standards-grounded design context; and case study 3 draws on clinical diagnosis. The breadth of this coverage—across evaluated decisions, normative standards, and diagnostic inference—is itself a form of validation, demonstrating that the framework is not domain-specific but rather captures structural features common to a wide class of expert judgment problems.

The three case studies differ in their empirical character, and readers are encouraged to calibrate their expectations accordingly. Case study 1, presented in Section 4.1, is the most formally developed and constitutes the core empirical contribution of the paper. It introduces specific Boolean function families as the representational substrate for expert reasoning, constructs confusion matrices over those families, and derives explicit hallucination error bounds. These derivations in Section 4.1.8 are quantitative and formal, and they are intended to demonstrate that the framework's claims about reasoning quality are not merely qualitative but can be given precise analytical expression. The use of Perplexity AI as a cognitive aide in constructing and interrogating these structures warrants a brief methodological note. Perplexity AI's multi-model architecture means that prompt-response dialogues were processed through several large language models simultaneously rather than through a single identifiable model. Per-model attribution is consequently not available. We do not treat this as a meaningful limitation, because the role of the LLM in this work is not classificatory. We are not using a language model as a judge, an annotator, or a component of the aggregation procedure itself; we are using it as a cognitive aide for eliciting, organizing, and stress-testing expert reasoning structures. The identity of the underlying LLM model is therefore not material to the validity of the results.

Case study 3 differs from case study 1 in an important respect. Rather than presenting new primary data, it reviews and reanalyzes experimental results from published studies that employed the same monotone function methodology. This case study demonstrates that the framework is consistent with, and can organize, empirical results that already exist in the literature on clinical diagnosis. Using established published findings as the evidential base allows the case study to connect the formal apparatus to a domain with well-characterized data. The presentation is framed explicitly as a reanalysis and synthesis of prior work, with original sources identified and the reanalysis distinguished clearly from the underlying experimental contributions of those studies.

## Case Study 1: Research Proposal Submission Decision

We used Perplexity AI [Perplexity, 2025] to determine whether it could generate useful questions for decision-making about responding to requests for proposals—a task typically performed by a team. Perplexity AI offers two key advantages for our study. First, it leverages **multiple LLMs** working in tandem, including GPT-4, Mistral Large, Claude 3, and proprietary models, allowing us to indirectly evaluate the capabilities of several LLMs simultaneously. Second, it **accesses the web in real time** to retrieve current information, overcoming the limitation of static knowledge cutoffs present in many other LLMs. Using web scraping and parsing techniques, it extracts relevant content from webpages and feeds this information to the LLMs.

This subsection traces a five-step prompt-response sequence designed to evaluate how well LLMs (Perplexity AI) can support a group of researchers assessing their readiness to respond to a funding call. The sequence begins with question generation and ends with the problem of aggregating individual answers into a coherent group judgment, progressively exposing the limitations of LLM-based decision support at each step.

### Proposing Questions

*Prompt.* A goal of a group of people is to evaluate their standing to respond to a call for proposals from a funding agency. They want to produce a list of Yes/No questions that each of them will answer before the meeting. Please propose questions.

*LLM Response.* Here is a list of Yes/No questions that a group could use to evaluate their readiness to respond to a call for proposals from a funding agency:

Have you thoroughly read and understood the call for proposals?

Does the proposed project align with the funding agency's priorities and goals?
Do you meet all the eligibility criteria specified in the call?
Is your organization legally qualified to receive and manage the grant funds?
Do you have the necessary expertise within your team to execute the proposed project?
Can you meet the proposal submission deadline?
Do you have the capacity to complete the project within the specified timeframe?
Is the proposed budget within the funding limits set by the agency?
Can your organization provide any required matching funds or in-kind contributions?
Do you have access to all the resources (e.g., facilities, equipment) needed for the project?
Have you identified potential partners or collaborators if required by the call?
Do you have experience managing similar grants or projects?
Can you fulfill all reporting and compliance requirements specified in the call?
Do you have letters of support or commitment from key stakeholders?
Have you reviewed, and can you adhere to the agency's data management and sharing policies?
Is your proposed project innovative or unique in its approach?
Can you demonstrate the potential impact and significance of your project?
Do you have a clear plan for disseminating the results of your project?
Have you identified potential risks and mitigation strategies for your project?
Are all team members available and committed to working on this proposal and project?

These questions cover various aspects of proposal readiness, including alignment with the funding agency's goals, team capacity, resource availability, and compliance with requirements. Answering these questions before the meeting can help the group assess their strengths and weaknesses in responding to the call for proposals.

## Proposing a Hierarchy of Questions *(Summary)*

*Prompt.* The team faces $2^{20} = 1{,}048{,}576$ possible scenarios, each defined by a unique combination of Yes/No answers to the 20 questions. To avoid evaluating all scenarios explicitly, the LLM was asked to propose a hierarchical grouping of the 20 questions, with no more than 5 child questions per node, where each node is either one of the original 20 questions or a generalization of them.

*Key Output.* The LLM produced a 7-group hierarchy (2 levels) organized around themes such as overall feasibility, eligibility, strategic fit, resources, collaboration, and compliance. Eleven generalized Yes/No questions replaced the original 20, and at each level a "No" answer triggered either an exit from the process or a recommendation for remedial action. Each expert should answer 11 questions and make a final decision based on the resulting 11-dimensional binary vector.

*Shortcoming.* The original 20 questions are never answered; they are absorbed into 11 generalized substitutes. The efficiency gain comes at the cost of transparency, since the individual questions that domain experts are best positioned to answer are no longer present in the evaluation.

## Preserving Actual Questions *(Summary)*

*Prompt.* The LLM was instructed that the original 20 questions cannot be replaced by generalized ones; generalized questions may only serve as gates that determine whether the team proceeds to answer the underlying original questions.

*Key Output.* The LLM revised the hierarchy so that 7 generalized questions function as entry conditions for 7 groups of original questions. All 20 original questions are retained as leaf nodes, and a final synthesis question asks for the overall recommendation based on all prior answers.

*Shortcoming.* The gate logic remains binary: a "No" at any generalized question blocks the team from engaging with its child questions entirely. This means a single negative answer at any level exits the process, which is too rigid for the graded and contextual judgments that proposal evaluation requires.

## Formulating Generalized Questions *(Summary)*

*Prompt.* The LLM was shown that a response pattern such as (Yes, No, Yes) for Q17, Q18, and Q19 should not automatically stop the proposal process, and was asked to redesign the hierarchy so that generalized nodes synthesize—rather than merely gate—the responses of their children.

*Key Output.* The LLM retained the same 7-level structure but added an explicit aggregation rule to each node. For example, the generalized question covering Q17–Q19 ("Is the project's potential impact sufficiently high and risks manageable?") was assigned the rule: answer Yes if at least two out of three child questions are Yes. The final decision node was similarly governed by a threshold: answer Yes if at least six out of seven generalized questions are Yes.

*Shortcoming.* Threshold-based rules of this kind—such as simple majority or fixed numeric cutoffs—require justification that the LLM does not supply. Domain experts may weigh questions asymmetrically, apply different thresholds from one context to another, or reject threshold logic altogether in favor of holistic judgment. The LLM provides no mechanism for eliciting or validating these preferences.

## Checking Accuracy of Generalized Answers

*Prompt.* How can we ensure that the generalized answer accurately reflects the responses to original questions like Q17, Q18, and Q19?

*LLM Response.* To ensure that the generalized answer accurately reflects the responses to Q17, Q18, and Q19, we need to establish a clear decision rule or aggregation logic that synthesizes the individual Yes/No answers into a single Yes/No outcome. This rule should reflect the relative importance of each question and account for the possibility that a single "No" does not necessarily invalidate the proposal.

*Step 1: Define the Purpose of the Generalized Question.* The generalized question for Q17, Q18, and Q19 is: "Is the project's potential impact sufficiently high and risks manageable?" This question aims to evaluate whether the combined responses to these three questions provide enough confidence to move forward with the proposal.

*Step 2: Establish Aggregation Logic.* Three approaches can aggregate the answers.

Under the Majority Rule, if at least two out of three answers are "Yes," the generalized answer is "Yes." This approach assigns equal weight to all three questions and prevents a single "No" from disproportionately affecting the outcome.

Under the Weighted Rule, each question is assigned a weight reflecting its relative importance—for example, Q17 (Impact) at 50%, Q18 (Dissemination Plan) at 25%, and Q19 (Risk Management) at 25%. The weighted score is then:

$$\text{Score} = \mathbf{1}[Q_{17} = \text{Yes}] \times 0.5 \; + \; \mathbf{1}[Q_{18} = \text{Yes}] \times 0.25 \; + \; \mathbf{1}[Q_{19} = \text{Yes}] \times 0.25$$

If Score ≥ 0.5, the generalized answer is "Yes." This approach reflects the judgment that demonstrating impact (Q17) is more critical than dissemination planning or risk management.

Under the Critical Threshold Rule, certain questions are designated as deal-breakers: a "No" on Q17 (low impact) automatically yields a generalized "No" regardless of the other answers, while a "Yes" on Q17 triggers majority-rule evaluation of Q18 and Q19.

*Step 3: Apply Aggregation Logic to Example Responses.* For Q17 = Yes, Q18 = No, Q19 = Yes, all three methods converge on the same outcome. Under Majority Rule, two out of three are "Yes," so the generalized answer is Yes. Under the Weighted Rule, Score = $0.5 + 0 + 0.25 = 0.75 \geq 0.5$, so the generalized answer is Yes. Under the Critical Threshold Rule, Q17 = Yes admits the remaining questions; one "Yes" out of Q18 and Q19 is sufficient, so the generalized answer is Yes.

*Step 4: Document and Communicate the Decision Rule.* To ensure consistency, the chosen aggregation logic should be defined clearly and shared with all team members before evaluation begins. A structured template or scoring system should record individual answers and calculate the generalized outcome transparently.

*Step 5: Validate the Generalized Answer.* After applying the aggregation logic, the team should review how well the generalized answer aligns with the individual responses, discuss edge cases, and refine weights or thresholds as needed to better reflect collective priorities.

By using one of these methods—Majority Rule, Weighted Rule, or Critical Threshold Rule—the generalized answer can be made to reflect the combined responses to Q17, Q18, and Q19 in a fair and transparent way. The same approach can be applied to all other nodes in the hierarchy.

## Consolidated Analysis: Sections 4.1.1.-4.1.5

The five-step exchange above reveals a consistent and instructive pattern: the LLM performs well at generating and organizing content but encounters **fundamental difficulties** when that content must **faithfully encode expert judgment** in a formally valid way.

At the question-generation stage (Section 4.1.1), the LLM produced a reasonable and broadly applicable set of 20 Yes/No questions covering alignment, eligibility, capacity, resources, compliance, and impact. The questions are competent, but they remain generic. Several require substantive analysis and domain expertise to answer, and even those that appear straightforward can conceal important nuance. Question 11, which asks whether potential partners have been identified, illustrates this well: a team may have identified partners but have no established relationship with them, which introduces a materially different risk profile than the question captures. This gap points to a practical conclusion: expert review and adjustment of LLM-generated questions is essential. The process could be supported by a curated question bank with software that allows practitioners to select and refine questions appropriate to their specific proposal context.

The hierarchy-building steps (Sections 4.1.2–4.1.4) demonstrate a second capability and a second set of limitations. The LLM successfully organized 20 questions into a multi-level structure that reduces cognitive load and allows early pruning of unfeasible paths. The resulting hierarchy is coherent and thematically well-motivated. However, three successive limitations emerged across these steps, each requiring a corrective prompt to address. Initially, the LLM replaced the original questions with generalized substitutes rather than preserving them as items to be answered. When instructed to preserve the originals, it adopted strict gate logic under which a single "No" at a generalized node blocked all its children—a structure too brittle for the nuanced, context-sensitive judgments experts actually make. When pushed further to soften the gates into synthesis rules, the LLM defaulted to fixed numeric thresholds (e.g., “Yes if at least 2 out of 3 are Yes”), which require justification the LLM does not provide, and which domain experts may reject entirely.

The aggregation problem explored in Section 4.1.5 is where these limitations become most consequential. Presented with the question of how to ensure a generalized answer accurately reflects its constituent responses, the LLM offered three options—Majority Rule, Weighted Rule, and Critical Threshold Rule—and demonstrated that all three yield the same answer for the specific example (Yes, No, Yes). This convergence on a single example obscures the fact that the three rules embody meaningfully different assumptions about the relative importance of questions and the conditions under which a “No” should be decisive. At the mathematical level, the Majority Rule corresponds to the simple condition $n(\text{yes}) > n(\text{no}) \Rightarrow \text{yes}$, where $n(\text{yes})$ and $n(\text{no})$ are the counts of positive and negative answers respectively—a special case of a far richer space of aggregation operators documented in the decision-support literature [Dujmovic, 2018; Beliakov et al., 2020]. The LLM has no mechanism for helping a domain expert determine which rule best matches their actual reasoning, and it offers no path forward if the expert rejects all three options presented.

In conclusion, this experiment demonstrates that the LLM can produce meaningful sets of questions, meaningful generalized questions, and a meaningful question hierarchy. The aggregation step, however, cannot be adequately handled by simplified arithmetic formulas of the kind the LLM generates. More sophisticated methods are needed—in particular, the approach based on the monotonicity principle presented in Section 3.1.1, which allows an expert’s preferences to be captured and validated without requiring the expert to accept any pre-specified formula.

## Model Specification Construction from First Principles

Determining whether to respond to a request for proposal (RFP) from a grant-awarding organization requires careful thought, especially for a small business with limited resources. For example, the award amount may be incompatible with the time and effort required to prepare a coherent and compliant response, due in part to the diversion of resources from other tasks. Moreover, the proposal may describe a problem for which the business may not have the requisite expertise to address fully, meaning the need for partnership with others external to the business. Thus, committing to a response is an investment that requires decision-making under uncertainty, informed by experience and circumstances.

Decision-making can be formalized as an explanatory process by systematically identifying the factors, their valid values, and the levels of organization that contribute to the decision. The process begins with a clear and concise description of the decision to be made—the root node of a tree whose valid values are ordinal and $k$-valued. From that root, the practitioner iteratively identifies factors that explain or support the parent node, creating a new child node for each such factor and specifying its valid values in turn. This expansion continues until every open node has been examined and no further explanatory factors can be identified, at which point the tree is complete.

The criterion for terminating expansion at any given node is not as straightforward to adjudicate as it may seem. Each node in the tree is a construct—an abstraction—and each factor applied to a parent increases the specificity and level of organizational explanation. One could argue, philosophically, that no end exists to the degree of possible explanation. However, practical limits help make problem-solving expedient, tractable, and explainable. For the kind of model specification construction considered here, that limit occurs when a factor is known or perceived to be observable, calculable, or computable. Such a factor becomes a leaf in the tree. Ultimately, that leaf will be tied to something that can provide the requisite

input—cast as a question to the user of the model learned from the specification, or as output from a measurement device, sensor, computer-based function, or intelligent agent.

```
{
"Should we respond to the RFP?": {
"What is the strength of our history with the Technical Monitor (TM)?": {
"Do we know the TM?": {},
"Have we worked with the TM?": {},
"Have we interacted with the TM in the last year?": {},
"Would we work with the TM again?": {}
},
"What is the strength of our history with the customer/sponsor?": {
"Do we have related work with the customer?": {},
"Have we interacted with the customer in the last year?": {},
"Would we work with the customer again?": {}
},
"Does the topic of (or do topics within) the RFP align?": {
"With our capabilities?": {},
"With our strategic business objectives?": {},
"With our prior work?": {}
},
"Do we have personnel with qualifications?": {
"To write the proposal?": {},
"To be key personnel?": {}
},
"Can we earn a fee?": {}
}
}
```

*Figure 2. Document in JavaScript Object Notation (JSON) representing a model specification for answering the question "Should we respond to the RFP (Request for Proposal)?". The specification is a tree with one root node, four branch nodes, and thirteen leaf nodes. For simplicity, each node is assumed to have a binary (Boolean) outcome: "no" (zero) and "yes" (one), in that order. However, nothing precludes an outcome from being one of a set of ordinal k-valued values (such as may be found in a Likert-type scale).*

Figure 2 provides one example of a model specification for the driving question "Should we respond to the RFP?" One could argue that the specification is incomplete and should be wider and deeper, with many more levels of organization. Indeed, model specification construction is a process of iterative refinement. If the answer to a question like "how do you make a decision about X?" is "it depends," then the objective is to tease out those dependencies in as great a specificity as practicable, with self-imposed limits to make model learning tractable. If the monotonicity principle holds, each level within the hierarchy will inform or support the one above it.

For clarity, the model specification is not a model. The specification *defines the parameters* by which a model may be learned. The model will manifest itself as an equation, formula, or other mathematical expression—relating an outcome (the parent node assigned a designated valid value) to a formula in disjunctive normal form (ORs of ANDs of factor-value combinations). A single specification could support multiple models. After all, the specification is an opinionated representation of a decision-making process, and so are the models learned from it.

The model comprises a specific combination of factors and values, and its utility lies in application to new situations—for instance, to the next RFP the organization encounters. Multiple experts could each have their own models that lead to the same root decision point. In each case, results can be explained in natural language, because every node in the tree is defined by a construct with valid values also expressed in natural language. Explainability is accordingly built into the structure. Coupled with the support that comes from answering questions posed at the leaf nodes, this produces the essential ingredients of accountable decision support.

One way in which this approach to model specification construction differs from the LLM-based techniques discussed above is that LLMs currently do not have the requisite degree of practical or applied knowledge to operationalize a leaf node. In fairness, the human expert creating the model specification may not have such knowledge either, due to gaps in awareness of where the relevant information resides and the mechanics of extracting it. However, in creating the specification, the expert asserts some degree of domain awareness and understanding. Moreover, owing to the abstractness and semantics of each factor in the hierarchy, the expert is presumably better positioned than the LLM to identify the appropriate explanatory level of organization—addressing the tradeoffs between specificity, tractability, and explainability. Although this difference may not persist given the current pace of technological advancement, to the extent that human decision-making still matters in human affairs, the approach presented here provides a way to encode and record tacit knowledge that is both tractable and explainable.

## Measuring LLM Hallucination in Boolean Proposal Functions

In general, large language models tasked with generating decision rules for complex scenarios—such as submit/not-submit Boolean functions for grant proposals—lack access to any team's true internal criteria. Any proposed function is thus a *hallucination*: a reasoned guess based solely on general knowledge of proposal processes. To quantify the **range of hallucination errors**, we construct an ordered set of **pure Boolean functions** (conservative, baseline, risky) over 5 and 20 binary attributes (dimensions, D), evaluate their exhaustive disagreement patterns, and compare one against another selectively—as either a "ground truth" team function or LLM-generated function—to better understand the effects of LLM hallucination on classification tasks. The three functions are explicitly ordered by strictness: $f_{cons}$ is the most demanding, $f_{risky}$ the least. This means every positive of $f_{cons}$ is also a positive of $f_{base}$, and every positive of $f_{base}$ is also a positive of $f_{risky}$—that is, $\{f_{cons} = 1\} \subseteq \{f_{base} = 1\} \subseteq \{f_{risky} = 1\}$, which also guarantees that the baseline requirement is met whenever the conservative requirement is met, for example.

**5-D attributes and functions**. Attributes: $x_1$(new idea), $x_2$(familiar topic), $x_3$(collaborators), $x_4$(qualified team), $x_5$(past success). 5-D functions with all $2^5 = 32$ input vectors defined are presented below.

**Conservative** function (all 5 variables required): $f_{cons}(\mathbf{x}) = x_1 \wedge x_2 \wedge x_3 \wedge x_4 \wedge x_5$. Only one vector (1,1,1,1,1) has $f_{cons}(1,1,1,1,1) = 1$ (3.1% of 32 vectors).

**Baseline** function (moderate): $f_{base}(\mathbf{x}) = x_1 \wedge x_4 \wedge [(x_2 \wedge x_3) \vee (x_2 \wedge x_5) \vee (x_3 \wedge x_5)]$. Only 4 vectors (12.5% of 32 vectors) have $f_{base}(\mathbf{x}) = 1$: (1,1,1,1,1), (1,1,1,1,0), (1,1,0,1,1), (1,0,1,1,1). The motivation for this function is that $x_1 \wedge x_4$ captures hard constraints (new idea $x_1$ and qualified team $x_4$) and $(x_2 \wedge x_3) \vee (x_2 \wedge x_5) \vee (x_3 \wedge x_5)$ capture 3 possible ways to get at least 2 of 3 soft (not-strictly-required) attributes (familiar topic $x_2$, collaborators $x_3$, past success $x_5$).

**Risky** function (new idea + at least 1 of 4 other attributes): $f_{risky}(\mathbf{x}) = x_1 \wedge (x_2 \vee x_3 \vee x_4 \vee x_5)$. The number of vectors with $f_{risky}(\mathbf{x}) = 1$ is 15 (46.9% of 32 vectors), all vectors with $x_1 = 1$ except (1,0,0,0,0).

**20-D attributes and functions**. For the 20 attributes listed in Section 4.1.1, the structure mirrors 5-D:

**Conservative**: 14 hard attributes and 4 of 6 soft attributes produce 22 vectors (see equation below); $f_{cons}(\mathbf{x}) = 1$ (0.0021%).

$$f_{cons}(\boldsymbol{x}) = \underbrace{\left(\bigwedge_{i=1}^{14} x_i\right)}_{\text{14 hard constraints}} \wedge \underbrace{\left(\sum_{i=15}^{20} x_i \geq 4\right)}_{\text{soft threshold on 6 attributes}} = 22$$

*Explanation:* The 14 required attributes are fixed: they must all equal 1, contributing exactly 1 combination. The variation comes entirely from the 6 soft attributes, where we count binary vectors of length 6 having at least 4 ones: $\binom{6}{4} + \binom{6}{5} + \binom{6}{6} = 15 + 6 + 1 = 22$. So, the total number of 20-D vectors in this case is $1 \times 22 = 22$.

**Baseline**: 12 hard and 3 out of 8 soft attributes produce 219; $f_{base}(\mathbf{x}) = 1$ (0.0209%).

**Risky**: 7 hard and 1 out of 13 soft attributes produce 8,191; $f_{risky}(\mathbf{x}) = 1$ (0.781%).

Percentages are computed relative to all $2^{20} = 1{,}048{,}576$ Boolean vectors yielding a decision to submit (that is, $f_{cons}(\mathbf{x}) = f_{base}(\mathbf{x}) = f_{risky}(\mathbf{x}) = 1$).

**Definition of hallucination error bounds**. LLM hallucination $H$ is defined as the *total number of misclassifications* when the true team function is not equal to the LLM guess. All functions use pure Boolean connectives (∧, ∨). Confusion matrices below show systematic patterns: the conservative function suffers from false negatives; the risky function suffers from false positives only. Formally, let $T: \{0,1\}^n \to \{0,1\}$ denote *the true (unknown) team Boolean function* and $L: \{0,1\}^n \to \{0,1\}$ denote the *LLM-generated (hallucinated) function*, where $n$ is the number of binary attributes (5 or 20).

**LLM hallucination error** is defined as:

$$H(T, L) = \sum_{\mathbf{x} \in \{0,1\}^n} \mathbb{I}[T(\mathbf{x}) \neq L(\mathbf{x})]$$

where $\mathbb{I}[\cdot]$ is the indicator function (1 if true, 0 if false).

**Normalized hallucination rate** (population **error rate**):

$$\boldsymbol{h}(T, L) = \frac{H(T, L)}{2^n}$$

where $h(T, L)$ is a proportion (0 to 1), specifically the population error rate—the fraction of all $2^n$ input vectors where $T \neq L$. To express percentage error rate, we use $100 \times h(T, L)$.

Example values:

$H(f_{\text{base}}, f_{\text{cons}}) = 197$, so $h(f_{\text{base}}, f_{\text{cons}}) = 197/1{,}048{,}576 \approx 0.000188$ (0.0188%)

$H(f_{\text{risky}}, f_{\text{base}}) = 7{,}972$, so $h(f_{\text{risky}}, f_{\text{base}}) = 7{,}972/1{,}048{,}576 \approx 0.00760$ (0.760%)

**Conditional error rates:**

On true positives: $h_1(T, L) = \frac{H_1(T,L)}{|\{x:T(x)=1\}|}$

On true negatives: $h_0(T, L) = \frac{H_0(T,L)}{|\{x:T(x)=0\}|}$

Whereas $H(T, L)$ is the population error rate (averaged over the entire input space), $h_1$ and $h_0$ are class-conditional error rates, which formally quantify LLM hallucination as the **symmetric difference** between the true team decision boundary and the LLM's guessed boundary in the full Boolean hypercube.

**Estimates of hallucination error bounds**. Below we present estimates of LLM hallucination errors with alternative true team functions.

**Case 1:** *True team is Conservative, LLM guesses Baseline*. In this case, the LLM over-submits without missing true positives but with adding extras.

| Dimension | Total Errors | Errors on true=1 | Errors on true=0 | Error Rate |
|---|---|---|---|---|
| 5-D | 3/32 | 0/1 (0%) | 3/31 (**9.7%)** | **9.4%** |
| 20-D | 197/1M | 0/22 (0%) | 197/1M (0.019%) | 0.019% |

In 5-D, the "Errors on true=0" are $4 - 1 = 3$ (the baseline positives not shared with conservative), out of $32 - 1 = 31$ true negatives. Similarly, in 20-D, the "Errors on true=0" are $219 - 22 = 197$ out of $1{,}048{,}576 - 22$ true negatives.

**Case 2:** *True team is Risky, LLM guesses Baseline.* In this case, the LLM under-submits, catching all true positives but missing cases the true function would approve.

| Dimension | Total Errors | Errors on true=1 | Errors on true=0 | Error Rate |
|---|---|---|---|---|
| 5-D | 11/32 | 11/15 (**73.3%**) | 0/17 (0%) | 34.4% |
| 20-D | 7,972/1M | 7,972/8,191 (**97.3%**) | 0/1M (0%) | 0.760% |

**Case 3:** *True team is Risky, LLM guesses Conservative*. In this case, maximum deviation occurs.

| Dimension | Total Errors | Errors on true=1 | Errors on true=0 | Error Rate |
|---|---|---|---|---|
| 5-D | 14/32 | 14/15 (**93.3%**) | 0/17 (0%) | **43.8%** |
| 20-D | 8,169/1M | 8,169/8,191 (**99.7%**) | 0/1M (0%) | 0.779% |

**Upper bounds**. If the true team function lies between the conservative and baseline functions in terms of submission aggressiveness, then the hallucination error when the LLM guesses the baseline function will be less than or equal to the case 1 error (3 misclassifications in 5-D; 197 misclassifications in 20-D). Similarly, if the true team function lies between the baseline and risky functions, then the hallucination error when the LLM guesses the baseline function will be less than or equal to the case 2 error (11 in 5-D; 7,972 misclassifications in 20-D).

**Key Insights**

*Unidirectional errors*: Conservative hallucinations produce false negatives (*missed submissions*); risky hallucinations produce false positives (*phantom submissions*).

Unidirectionality is a direct consequence of the nesting structure of the conservative, baseline, and risky functions, making errors between them unidirectional:

When the LLM guesses a looser function than the truth (cases where the LLM is more conservative)—fails to flag vectors the true function would approve—**only false negatives** occur.

When the LLM guesses a stricter function than the truth (cases where the LLM is more permissive)—approves vectors the true function would reject—**only false positives** occur.

This explains why every confusion matrix in the Cases above has a zero in either the "Errors on true=1" or "Errors on true=0" cells.

*Dimensionality effect:* Low-D (5-D) errors are in the range of ~10–44%; high-D (20-D) errors are <1% due to *sparsity* (positives occupy most of the $2^n$ space).

*Upper bounds:* If the true function is between conservative and baseline functions, then its error ≤ case 1 error. If it is between baseline and risky functions, then its error ≤ case 2 error.

*Critical conditional errors:* While population error rates shrink with dimension, the *conditional error rate on true positives approaches* ***100%*** *for bad guesses* (e.g., conservative guessing risky positives).

This framework bounds LLM hallucination when guessing opaque team decision rules. **Practical implication**: In sparse high-D spaces, absolute disagreements look large (8k vectors), but population error rates stay tiny (<1%)—yet *rare true positives* remain catastrophically misclassified by mismatched guesses.

## Case Study 2: Cybersecurity System Design

### Motivation and Approach

Fundamentally, designing a new engineering system requires expert input that is unavailable to an LLM to help develop such a system. Designing and maintaining a cybersecurity system for a large organization is one example we consider in this section. The effort requires domain expertise, like identifying categories of information that need to be protected and judging the security impact associated with each of them as low, medium, or high. The effort also requires that a domain expert define, analyze, and evaluate numerous combinations of risk factors and their mitigations—judgments that could be difficult under ordinary circumstances, but that uncertainty exacerbates.

For system resilience, the evaluation and mitigation of risks is not a singular activity but a process. The process must be revisited as system use and external threats challenge assumptions about the system in operation. For an **expert mental model** to have utility beyond the reason for its original creation, to avoid having to re-enlist a domain expert in its update, forethought must be given to the variety of factors that impinge on the challenge the model addresses. For perspective, a mental model with just 5 factors will require $3^5 = 243$ elicitations of an expert if each factor has only 3 answer options (e.g., low, medium, high). Similarly, with 8 factors and 3 values for each, the total space contains $3^8 = 6561$ alternatives. As the number of factors increases, so does the value of framing challenges to respect the monotonicity principle.

The main benefit of using monotone Boolean or ordinal *k*-valued functions is in **cutting out the number of questions** needed to ask the domain expert to build the expert's mental model. The major difference of this cybersecurity task from the research proposal decision task considered above is that the set of cybersecurity questions that need to be addressed by domain experts can be taken directly from a standard [National Institute of Standards and Technology, 2004].

## Cybersecurity Questions and Answers Formalized

In the US, the Federal Information Security Management Act (FISMA) standardizes and consolidates security review and reporting across government agencies. It requires that agencies complete the authorization to operate (ATO) process [U.S. General Services Administration] before using, buying, or building software to minimize and manage risk. Below we present a subset of security impact questions at two levels as defined in the standard [National Institute of Standards and Technology, 2004], wherein the questions are already organized to respect the monotonicity principle, and formalize it with the introduction of mathematical notation:

| *Factor* | *Description* |
|---|---|
| $x_1$ | **How confidential is the system's data?** |
| $x_{11}$ | Are there secrets or private information that you need to protect? |
| $x_{12}$ | Is there personal identifiable information (PII), contract data, or other special kinds of data? |
| $x_2$ | **What is the importance of the system's integrity?** |
| $x_{21}$ | What would be the impacts of the system getting defaced? |
| $x_{22}$ | What could happen if the data was altered? |
| $x_3$ | **How important is the availability of the data?** |

Let $x_1, x_2,$ and $x_3$ be answers for the upper-level questions 1-3, $x_{ij}$ for the respective sub-questions. For each question, the standard [National Institute of Standards and Technology, 2004] establishes the security impact level by the highest answer to the three questions (FISMA low, FISMA moderate, or FISMA high, which we encode as 1, 2, and 3, respectively). Respectively, we define 3-valued functions $F_1, G_1,$ and $G_2$:

$$F_1(x_1, x_2, x_3) = k(k = 1, 2, 3)$$

$$G_1(x_{11}, x_{12}) = x_1(x_1 = 1, 2, 3)$$

$$G_2(x_{21}, x_{22}) = x_2(x_2 = 1, 2, 3)$$

We combine these functions as follows:

$$F_1(G_1(x_{11}, x_{12}), G_2(x_{21}, x_{22}), x_3) = k(k = 1,2,3)$$

So, for example, if $G_1(1,2) = 2$, $G_2(3,1) = 3$, and $x_3 = 2$, we have

$$F_1(G_1(1,2), G_2(3,1), 2) = F_1(2,3,2) = \max(2,3,2) = 3 \text{ [FISMA high]}$$

The standard defines the value of integration by a simple **maximum (max)** formula, which differs from the aggregation rules presented earlier in Section 3.2 and confirms our concern that the set of rules suggested by the LLM in Section 3.2 is limited. Also, the maximum formula represents the worst-case security approach, where out of several impact factors the impact of the most severe factor is assigned to the set of these factors. Although this approach seems justified for a high-risk application, it may not represent what the domain expert thinks in some situations. The standard enforces the worst-case approach. The role of the domain expert is only to provide answers for the lower-level questions, because answers to the upper-level questions are automatically generated by the max aggregation formula.

The standard provides a useful starting point for organizations not obliged to use it, but not using the max formula as the aggregation rule has consequences. For $G_1(x_{11}, x_{12})$, we have 3 values of $x_{11}$ and 3 values of $x_{12}$ representing 9 combinations. Each of these 9 combinations can have 3 values of $x_1$ for a total 27 combinations. In the same way, $G_2(x_{21}, x_{22})$ will have 27 combinations. Thus, $G_1(x_{11}, x_{12})$, $G_2(x_{21}, x_{22})$, and $x_3$ will have $27 \times 27 \times 3 = 2187$ combinations total. In other words, the expert mental model at this level will require 2,187 answers. The value of $k$ of $F_1$ also has 3 values. Thus, the dimension of the space of the total function $F_1(G_1(x_{11}, x_{12}), G_2(x_{21}, x_{22}), x_3) = k$ is $2187 \times 3 = 6561$. Recall that the use of the max formula allows a domain expert to answer only $3^5 = 243$ questions for producing the full model at this level.

Given this combinatorial explosion, the key issue in constructing an expert mental model is arranging a dialog with a domain expert and LLM to get 5 answers to independent factors $x_{11}, x_{12}, x_{21}, x_{22}$, and $x_3$.

Consider an example of questions where the sample answers appear after the colon (:) and factors such as $y_1$ have natural language descriptions:

*Please list secrets and private information that need to be protected*: A, B, C

*Please list factors that define security impact of item A today*: $y_1, y_2$

*Please list factors that define security impact of item B today*: $y_3, y_4$

*Please list factors that define security impact of item C today*: $y_5, y_6, y_7, y_8$

*What is the likely presence of factor $y_1$ today (low, medium, high) for item A?*: medium

*What is the likely presence of factor $y_2$ today (low, medium, high) for item A?*: low

*What is the likely presence of factor $y_3$ today (low, medium, high) for item B?*: high

*What is the likely presence of factor $y_4$ today (low, medium, high) for item B?*: medium

*What is the likely presence of factor $y_5$ today (low, medium, high) for item C?*: low

*What is the likely presence of factor $y_6$ today (low, medium, high) for item C?*: low

*What is the likely presence of factor $y_7$ today (low, medium, high) for item C?*: high

*What is the likely presence of factor $y_8$ today (low, medium, high) for item C?*: high

*What is the security impact of medium $y_1$ and low $y_2$ on item A?*: medium

*What is the security impact of high $y_3$ and medium $y_4$ on item B?*: medium

*What is the security impact of low $y_5$ and $y_6$ and high $y_7$ and $y_8$ on item C?*: high

*What is the security impact on A, B, and C when the impact on A and B is medium and the impact on C is high?*: high

Formalizing the outcomes, we have security questions *S* with answers for security impact of factors *A*, *B*, and *C* at the current time:

$$S_A(y_1, y_2) = S_A(\text{medium}, \text{low}) = S_A(2,1) = \text{medium} = 2$$

$$S_B(y_3, y_4) = S_B(\text{high}, \text{medium}) = S_A(3,2) = \text{medium} = 2$$

$$S_C(y_5, y_6, y_7, y_8) = S_C(\text{low}, \text{low}, \text{high}, \text{high}) = S_A(1,1,3,3) = \text{high} = 3$$

The total security impact $T$ of factors $A$, $B$, and $C$ may be described as follows:

$$T\left(S_A(y_1, y_2), S_B(y_3, y_4), S_C(y_5, y_6, y_7, y_8)\right) = T(2,2,3) = \max(2,2,3) = 3$$

Note that factors $A$, $B$, and $C$ were described as valid only at the current time (*today*). Building a resilient expert model would need to ask about many other possible values of variables and their combinations. The theory of monotone Boolean and ordinal $k$-valued functions permits a decrease in the number of these questions in the same way as we presented in Section 3.1.1 for the research proposal decision task. The difference is that Section 3.1.1 considered Boolean functions (with 2 values) whereas we now consider 3-valued functions. We presented factors $y_1, \ldots, y_8$ as abstractions above. The actual sources of these factors are not limited to domain experts. Doctrine (e.g., [National Institute of Standards and Technology, 2004; U.S. General Services Administration]; see Table 2) could be used directly or to prompt an LLM to generate candidate factors. (We presented an LLM-based approach in Section 4.1.)

*Table 2. Potential impact definitions for security objectives [National Institute of Standards and Technology, 2004].*

| | **POTENTIAL IMPACT** | | |
|---|---|---|---|
| **Security Objective** | **LOW** | **MODERATE** | **HIGH** |
| ***Confidentiality.*** Preserving authorized restrictions on information access and disclosure, including means for protecting personal privacy and proprietary information.<br>[44 U.S.C., SEC. 3542] | The unauthorized disclosure of information could be expected to have a **limited** adverse effect on organizational operations, organizational assets, or individuals. | The unauthorized disclosure of information could be expected to have a **serious** adverse effect on organizational operations, organizational assets, or individuals. | The unauthorized disclosure of information could be expected to have a **severe or catastrophic** adverse effect on organizational operations, organizational assets, or individuals. |
| ***Integrity.*** Guarding against improper information modification or destruction, and includes ensuring information non-repudiation and authenticity.<br>[44 U.S.C., SEC. 3542] | The unauthorized modification or destruction of information could be expected to have a **limited** adverse effect on organizational operations, organizational assets, or individuals. | The unauthorized modification or destruction of information could be expected to have a **serious** adverse effect on organizational operations, organizational assets, or individuals. | The unauthorized modification or destruction of information could be expected to have a **severe or catastrophic** adverse effect on organizational operations, organizational assets, or individuals. |
| ***Availability.*** Ensuring timely and reliable access to and use of information.<br>[44 U.S.C., SEC. 3542] | The disruption of access to or use of information or an information system could be expected to have a **limited** adverse effect on organizational operations, organizational assets, or individuals. | The disruption of access to or use of information or an information system could be expected to have a **serious** adverse effect on organizational operations, organizational assets, or individuals. | The disruption of access to or use of information or an information system could be expected to have a **severe or catastrophic** adverse effect on organizational operations, organizational assets, or individuals. |

Direct use of doctrine could take the form illustrated in Table 3, where we draw our example from placeholder factors *A*, *B*, and *C* above, this time with descriptive labels: "public information", "investigative information", and "administrative information", respectively, a different set of sub-factors: *confidentiality*, *integrity*, and *availability*, and their valid values: NOT APPLICABLE [NA] = 0, LOW = 1, MODERATE = 2, HIGH = 3.

*Table 3. Security Classification (SC) applied to information types [National Institute of Standards and Technology, 2004].*

| Security Classification | Scenario | Formulation |
|---|---|---|
| public information | {(**confidentiality**, NA), (**integrity**, MODERATE), (**availability**, MODERATE)} | $SC_{\text{pi}}(0,2,2)$ |
| investigative information | {(**confidentiality**, HIGH), (**integrity**, MODERATE), (**availability**, MODERATE)} | $SC_{\text{ii}}(3,2,2)$ |
| administrative information | {(**confidentiality**, LOW), (**integrity**, LOW), (**availability**, LOW)} | $SC_{\text{ai}}(1,1,1)$ |

## Case Study 3: Healthcare Decisions

The current state of the art in explainable AI and machine learning (XAI) in healthcare presents a fundamental challenge: how to put domain experts in meaningful control of AI and ML processes so that the outputs of high-stakes diagnostic and predictive models can be audited, corrected, and trusted. Addressing this challenge requires resolving two interdependent problems—giving clinicians tools to observe and discover patterns in high-dimensional data without loss of information and extracting and encoding expert models that can be compared and integrated with data-driven ones. The monotone $k$-valued function methodology developed in Section 3 directly targets both problems. This case study does not present new primary empirical data. Rather, it reviews and synthesizes experimental results from a series of published studies that applied this methodology to clinical diagnosis tasks, demonstrating that the formal framework is not only theoretically well-formed but has already been instantiated successfully in real healthcare settings. This review approach is appropriate for the scope of a framework and algorithm paper: the contribution here is to show that the machinery of Section 3 is consistent with, and capable of organizing, existing empirical findings across a range of diagnostic domains.

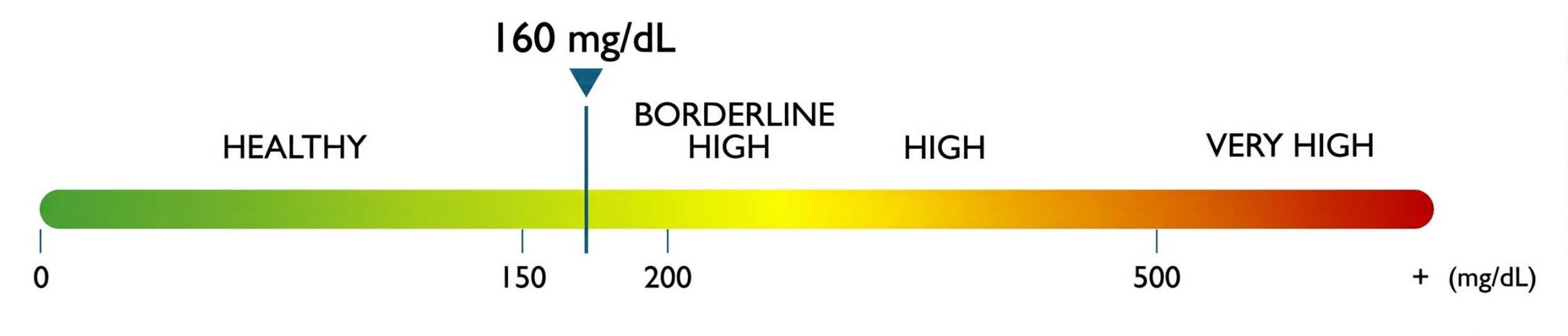

*Figure 3. Illustration of a continuous scale binned with qualitative labels. [Values taken from https://ada.com/biomarkers/triglycerides/].*

To see why this methodology is well-suited to clinical contexts, consider the structure of a typical diagnostic task. Suppose a clinician is interpreting results from 20 blood tests, each of which can be assigned one of 5 qualitative values—"significantly below normal," "slightly below normal," "normal," "slightly above normal," and "significantly above normal." These discrete values represent consecutive ordinal bins of continuous data, a standard accommodation that reduces the number of alternatives a clinician must hold in working memory. The reduction is necessary but costly: the number of possible factor combinations is still $5^{20} \approx 9.5 \times 10^{13}$, and the conversion of continuous data to qualitative categories introduces threshold uncertainty, since different clinicians may draw category boundaries differently or prefer fuzzy transitions between adjacent categories, as illustrated in Figure 3. The triglyceride biomarker provides a concrete illustration of this binning structure. According to standard clinical guidelines, triglyceride levels are classified into four qualitative categories: healthy (below 150 mg/dL), mildly high (150–199 mg/dL), high (200–499 mg/dL), and very high (above 500 mg/dL). In the $k$-valued function framework, this factor would be represented as a $k = 4$ input variable taking values in an ordered set, with monotone reasoning constraints encoding the clinical expectation that higher triglyceride levels are associated with greater diagnostic concern. A clinician's decision rule over a panel that includes this factor—for example, "flag for cardiovascular risk assessment if triglycerides are high or very high and at least two other lipid markers are elevated"—can be expressed as a monotone Boolean function over the factor space, making the rule formally representable, storable, and comparable across clinicians.

Only a small fraction of the possible factor combinations will appear in any recorded dataset, meaning that in practice clinicians apply a form of implicit aggregated logic, weighting critical factors and suppressing less relevant ones. Developing an explicit expert mental model in the form of a monotone function offers several advantages over leaving this logic implicit. It allows the model to be compared against empirical data and corrected when discrepancies arise. It supports comparison across clinicians to identify consensus and divergence. And it allows machine learning models trained on incomplete or biased data to be audited and corrected using the expert model as a reference standard [Kovalerchuk et al., 2012; Huber, Kovalerchuk, 2024].

Published studies applied exactly this methodology to three diagnostic domains: breast cancer, diabetes, and cardiovascular disease [Kovalerchuk et al., 1996; Kovalerchuk et al., 2001; Kovalerchuk et al., 2012]. Notably, the same 11-dimensional binary mammography feature space serves as a shared empirical testbed, allowing findings on elicitation efficiency, classification accuracy, and visual rule analysis to be read together:

**Elicitation efficiency.** The 1996 paper establishes the core quantitative result: interactive restoration of the two nested diagnostic functions—$f_1$ and $f_2$—required only 29 calls per function under independent search (Table 2, page 115; 13 in $E_5$, 12 for $\psi$, 4 for $\varphi$), 36 calls under the sequential strategy $\langle f_2, f_1 \rangle$ and 38 under joint search (Table 3, page 115)—representing 1,812 fewer queries than the Hansel upper bound of 1,848 and 4,058 fewer than exhaustive search. The full expert dialogue was completed in under an hour.

**Elicited function structure.** Both the 1996 and 2001 papers recover the same four functions: $\varphi(w_1, w_2, w_3) = w_2 \vee w_1 w_3$ and cancer suspicion) at the clinical level, with both sharing the common $\psi(y_1, \dots, y_5) = y_1 \vee y_2 \vee y_3 y_4 y_5$ at the sub-attribute level, and $f_1$(biopsy) and $f_2$(sub-formula $A = x_1 x_2 \vee x_3$ and the nesting constraint $f_2 \Rightarrow f_1$confirmed throughout.

**Classification accuracy.** The 2001 paper provides the quantitative benchmark comparisons. On a leave-one-out test across 156 cases, MMDR's (Machine Method for Discovering Regularities, data mining method) 18 most reliable rules (significance ≥ 0.95) achieved 96.6% accuracy with only 3 false positives, while the neural network collapsed from 100% training accuracy to 66%, LDA (Linear Discriminant Analysis) reached only 76%, and decision trees reached 76–82% on training data alone—all weaker than MMDR's harder generalization figure. The clinically decisive false-negative counts were 3–8 (MMDR), 8–9 (decision tree), 19 (LDA), and 26 (neural network).

**Expert–data-mining divergence and visualization.** The 2001 and 2012 papers together characterize where the two rule sources disagree. The statistically significant data-mining rule linking calcification density and volume to malignancy had no expert counterpart and was rated risky by the radiologist without additional evidence; the expert's analogous rule required a shape-irregularity condition the data alone could not confirm. In a direct reliability test on 30 cases, expert-extracted rules matched the radiologist's own diagnoses in 18 of 30; the remaining 12 required features outside the 11-attribute set, exposing a rep-

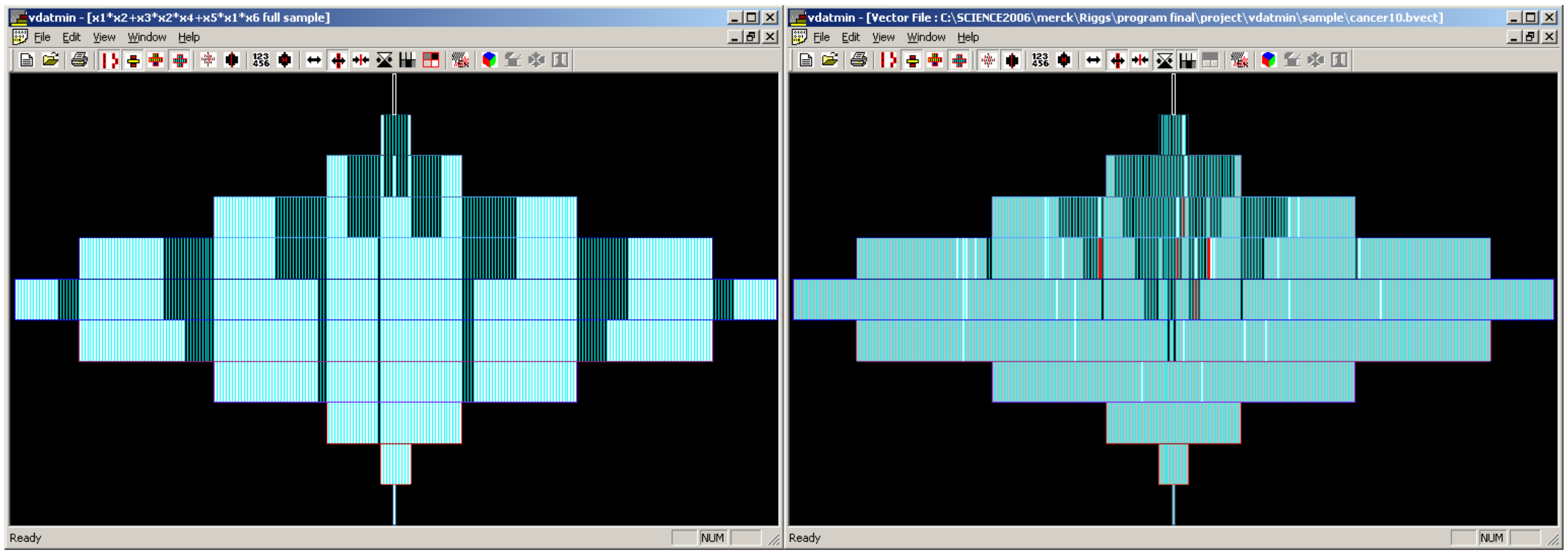

(a) rules extracted by a data mining algorithm.

(b) rules extracted from actual data (black and white) and monotonicity (light and dark gray); red indicates monotonicity violation.

*Figure 4. Visual comparison of rules [Kovalerchuk et al., 2012, Figure 9].*

resentation gap made detectable by the monotone function framework. The 2012 paper adds a visual dimension to this finding: an MDF (multiple disk form)/chain-based visualization revealed that the data-mining "final" rules contained a region (Figure 4) where the monotone expansion boundary fell below the data-mining boundary—meaning some cancer cases would have been misclassified as benign—while the expert rules remained fully nested and consistent throughout. Alternative parallel coordinate plots of the same data had failed to distinguish benign from malignant cases at all, making Figure 4 a visualization-specific positive empirical result.

Across these studies, the expert-in-the-loop approach combined expert knowledge and machine intelligence by involving clinicians in both the training and testing stages of algorithm development, rather than treating expert judgment as a one-time input to be superseded by the learned model. The consistency of these results across three distinct diagnostic studies—each with its own data characteristics, factor structure, and clinical reasoning conventions—provides empirical support for the generality of the framework. The present paper's contribution is to formalize and extend the methodological foundations underlying those studies, and to situate them within the broader aggregation and reasoning framework developed in Sections 2 and 3.

# Discussion

## Revisiting the Four Contributions Considering the Evidence

The four contributions claimed in Section 1.5 can be mapped to the evidence as follows. The first—a causal prompt engineering framework integrating EMMs with LLMs—is demonstrated across all three case studies, with Section 4.1 providing the most complete instantiation of the full pipeline, from LLM-assisted factor generation through expert-validated aggregation logic. The second—the four-step EMM construction algorithm exploiting monotone Boolean and $k$-valued functions—is given quantitative grounding in Section 3.1.1, where four expert elicitations classified 64.8% of the 128-scenario input space with a leverage ratio of nearly 20 automatic inferences per elicitation, and in Section 4.2, where the same principle applied to 3-valued functions eliminates the need to navigate up to 6,561 unconstrained combinations. The third—validation across three qualitatively distinct knowledge environments—is provided by the case studies collectively, with the important qualification that case study 3's contribution is prior confirmational rather than primary-evidentiary, as acknowledged in Section 4.3 and elaborated in Section 5.3. The fourth—formal quantification of LLM hallucination in tacit-knowledge settings—is developed analytically in Section 4.1.8, whose implications for the LLM-alone versus EMM-guided contrast are discussed in the section immediately following.

## Hallucination Analysis as a Comparison Between LLM-Alone and EMM-Guided Conditions

The analysis in Section 4.1.8 is most usefully understood as a controlled comparison between two conditions for supporting a binary proposal submission decision. In the LLM-alone condition, a large language model is asked to generate a decision function without access to the team's actual criteria—producing a hallucination in the precise sense this paper uses. In the EMM-guided condition, the elicitation procedure in Algorithm 1 constructs a decision function grounded in the expert's actual judgments, with the LLM serving only in its demonstrably competent supporting role of factor identification and hierarchy structuring. The conservative, baseline, and risky Boolean function families of Section 4.1.8 instantiate the realistic range of misclassification to which the LLM-alone condition is susceptible; the confusion matrices and error bounds derived there establish how large that range can be.

The practical implication for organizations follows directly from those results. Population-level error rates in high-dimensional spaces appear reassuringly small, but as Section 4.1.8 demonstrates, this figure is actively misleading: it reflects the sparsity of the positive region rather than the accuracy of the LLM as a decision guide. For any team whose actual decision criteria yield positive decisions on only a small fraction of the input space, the conditional error rate on true positives is the operationally consequential figure, and the analysis shows it can approach totality for plausible function mismatches. The EMM-guided condition addresses this not by calibrating the LLM's guess more carefully but by replacing the guess entirely: the elicitation builds the decision function from the expert's actual judgments, validated for internal consistency by the monotonicity lattice, and the resulting model converges to the true decision boundary through a dialogue whose efficiency is formally guaranteed.

Beyond this practical contrast, the analysis opens a category of empirical finding about LLMs that was not obtainable before. Prior hallucination research compares LLM outputs to known ground truths—benchmark answers, verified facts, and human annotations. In tacit-knowledge settings, no such external reference exists: the expert's decision function has never been written down, and no dataset can serve as the basis for direct measurement. The framework circumvents this by constructing the space of functions the LLM might plausibly hallucinate and bounding the disagreement analytically—yielding a finding that is directional and predictable rather than random, and that no benchmark comparison or retrieval-based evaluation could have uncovered. One limitation should be acknowledged: the three Boolean families are analytical constructions rather than empirically observed functions, so the analysis provides bounds rather than point estimates. The conditional error figures should be read as establishing the plausible range of harm for a given team rather than predicting an exact rate. Note that although a 5-dimensional experiment can be conducted with real domain experts, the full 20-dimensional experiment without a hierarchy is infeasible, as it would require an expert to evaluate over a million situations.

## The Proposed Framework in Relation to Prior Hybrid Approaches

The literature reviewed in Section 2 converges on the diagnosis that motivates this paper: LLMs encode vast domain knowledge and can approximate or exceed expert performance on narrow, well-specified tasks, but they do so through mechanisms categorically different from the grounded, causal, metacognitive structures that define genuine expertise. The productive response is hybrid architecture, coupling LLMs with explicit expert knowledge representations rather than asking either to substitute for the other. Table 4 extends the comparative survey of Section 2.7 by adding the proposed framework as a row, enabling a direct comparison across the six dimensions used to characterize prior work.

*Table 4. Comparative overview of representative works including the proposed framework.*

| Work | Methodology | Application Domain | Hallucination Mitigation | Tacit Knowledge Handling | Computational Complexity | Main Limitations |
|---|---|---|---|---|---|---|
| Sarma et al., 2026 | Expert decision tree + LLM prompting | Psychiatric diagnosis | Expert constraints suppress overdiagnosis | Decision tree encodes clinician heuristics | Low (inference-time only) | Static tree; not adaptive |
| Rollwage et al., 2026 | Cognitive layer architecture | Psychotherapy (CBT) | Structured clinical scaffolding | CBT protocols embedded in architecture | Moderate (additional layer) | Domain-specific; not generalizable |
| Wu et al., 2025 | ACT-R + LLM adapter (LLM-ACTR) | Manufacturing decisions | Cognitive grounding reduces drift | Expertise levels modeled explicitly | High (training required) | Narrow industrial domain |
| Bian, 2025; Ibrahim et al., 2024 | KG-augmented LLM | General / biomedical | KG grounding reduces factual errors | Structured ontologies encode domain knowledge | Moderate–High | KG construction costly; coverage gaps |
| Kong et al., 2024 | Role-play prompting | General reasoning | Indirect (domain framing) | Statistical approximation of expert discourse | Very low | Reflects training data, not structural reasoning |
| Martin-Boyle et al., 2026 | Expert schema elicitation | Scholarly QA | Error taxonomy guides evaluation | Meta-level model of LLM failure modes | Low | Descriptive; not preventive |
| Basu et al., 2026 | Causal reasoning augmentation | Planning under uncertainty | Causal grounding prevents confabulation | Psychologically grounded causal model | Moderate | Requires formal causal model specification |
| **Present paper** | **Monotone Boolean / *k*-valued function elicitation with LLM-assisted factor** | **Proposal review, cybersecurity design, clinical diagnosis (review)** | **Expert-grounded EMM replaces LLM-generated decision functions; formal hallucination error bounds derived** | **Optimized elicitation dialogue encodes knowledge absent from any accessible document; handles both fully tacit and** | **Low at inference time; moderate at elicitation phase, bounded by formal** | **Monotonicity requires careful question design; single-expert EMM may not generalize; healthcare contribution rests on prior work; no** |

| Work | Methodology | Application Domain | Hallucination Mitigation | Tacit Knowledge Handling | Computational Complexity | Main Limitations |
|---|---|---|---|---|---|---|
| | **hierarchy (Algorithm 1)** | | **over Boolean function families** | **doctrine-scaffolded environments** | **monotone efficiency guarantees** | **large-scale controlled study** |

Examining the table, two dimensions distinguish the proposed framework from prior work most sharply. First, on tacit knowledge handling: the approaches by Sarma et al., Rollwage et al., and Wu et al. all encode expert knowledge effectively, but in each case that knowledge must have been **previously articulated, protocolized** (as in CBT competencies), or represented in a cognitive architecture. Knowledge graph approaches and role-play prompting are bounded even more directly by what has been recorded or absorbed into training data. The present framework is the only entry that explicitly addresses the case in which critical knowledge has **never been written** down anywhere, providing a domain-general elicitation procedure for precisely this class of knowledge. Second, on the computational complexity and generalizability dimensions: the ACT-R integration requires significant training overhead and is tied to a specific cognitive architecture and domain; the cognitive layer architecture of Rollwage et al. was demonstrated only in psychotherapy. The present framework's elicitation phase requires expert time but not model **training**, and the case studies span three qualitatively different knowledge environments. For instance, in the proposal submission task, it is unrealistic to have sufficient training data applicable to the current solicitation and team configuration. The formal efficiency guarantee—that monotone inference will always reduce the number of required elicitations, with the reduction compounding as factor count grows—is a property of the mathematical theory rather than a domain-specific optimization, and it therefore transfers automatically to any decision domain for which the monotonicity is arranged in the process of generation of factors. Importantly, monotonicity is **not an assumption** but a result of factor design where a true value is matched with a positive output with LLM support.

## Generalizability Across Three Qualitatively Distinct Knowledge Environments

The three case studies span three fundamentally different relationships between accessible knowledge and the decision at hand, and examining them together reveals a consistent structural pattern that confirms the framework's generalizability.

Case study 1 represents the purest instance of the problem class the paper addresses: a decision in which critical knowledge is entirely tacit, personal to the team, and absent from any accessible document. No retrieval-augmented generation strategy, however well engineered, can supply what a team knows about its own relationship with a technical monitor, its realistic assessment of internal bandwidth, or its judgment about whether a particular funding call represents a genuine strategic fit. The LLM experiments in Section 4.1 confirm this directly and emphatically: Perplexity AI generated questions and a hierarchy that a domain expert would recognize as useful—the comparison in Section 4.1.7 shows that LLM-generated questions are broadly comparable to expert-generated ones—but it could not supply the aggregation logic that reflects a specific team's decision criteria. The three aggregation rules the LLM proposed are not wrong in the abstract; they are inadequate because they hallucinate an arbitrary decision procedure instead of the one the team actually uses, and the gap between those two things is precisely what the EMM framework exists to close. The framework's value in this environment is maximal because the LLM's contribution is bounded by the limits of **generic decision knowledge** not is not actual expert tacit domain knowledge.

Case study 2 represents a different epistemic configuration: one in which a normative document provides a partial scaffold for the factor hierarchy and the aggregation formula. Here the LLM can legitimately assist in extracting factor hierarchies from existing doctrine, and the FIPS 199 max formula is a valid and defensible aggregation rule for its designated scope. But the standard does not—and structurally cannot—specify which systems within a given organization hold which categories of data at what sensitivity level, nor what the operational impact of a compromise would be in that organization's specific context. That content must still be elicited from a domain expert. The automated mode of the framework handles the structural extraction from doctrine; the elicitation dialogue handles the **organization-specific content**. Case study 2 thus illustrates that the framework's value is not confined to fully tacit knowledge environments but extends naturally to environments where doctrine provides a starting scaffold, supplying precisely the portion of the decision logic that doctrine leaves underspecified.

Case study 3 extends the framework's reach to a domain in which prior empirical models, derived from data and expert mental models, have already been produced. The published studies reviewed in Section 4.3 applied the monotone *k*-valued function methodology to three clinical diagnostic domains and produced results that converge on a consistent finding: expert-extracted monotone functions, obtained through optimized elicitation dialogues, outperform purely data-driven models on the clinically consequential class of errors, are internally consistent in ways that data-mining rules need not be, and can be visualized in forms that reveal disagreements between expert reasoning and statistical pattern recognition that standard parallel coordinate plots cannot distinguish. The framework's contribution in this environment is connective: it provides the formal structure that allows expert and data-driven models to be placed in direct comparison, visualized, and corrected against each other. Even in a domain with well-characterized empirical data, the **expert's decision function adds information** that the data alone cannot supply.

The three cases together support a principle of consistent advantage across knowledge environments. Where knowledge is entirely tacit, the framework provides the team-specific decision logic that LLMs hallucinate when asked directly. Where knowledge is partially documented, the framework provides the organization-specific content that doctrine leaves underspecified. Where prior empirical models exist, the framework provides the formal structure that makes expert and data-driven models mutually legible, comparable, and correctable. In every case, the LLM is a genuinely useful participant—for generating candidate factors, structuring hierarchies, and stress-testing proposed aggregation rules—but it is not sufficient in any of them. The EMM framework is the complementary mechanism that converts LLM assistance from a source of plausible-but-potentially-hallucinated output into a principled component of an accountable decision process. This is not a trivial observation: the three knowledge environments span a meaningful portion of the space of real organizational decisions, and the consistency of the advantage across them is what justifies the framework's claim to generality.

## Limitations as Revealed by the Experiments

As acknowledged in Section 4.3, every quantitative result cited in the healthcare case study—elicitation efficiency figures, classification accuracy comparisons, visualization findings—comes from published work conducted by other groups of researchers in prior years. The methodological contribution of case study 3 is rather a synthesis and integration of multiple prior works, not evidentiary work in the primary sense. The framework has been validated in new domains in case studies 1 and 2, and it is consistent with and capable of organizing prior validation in healthcare, but it has not been subjected to new primary empirical testing in clinical settings. This matters because the three domains exhibit different characteristics—particularly in the relationship between expert knowledge and the available data—and it is not guaranteed that the efficiency results from the proposal submission domain (four elicitations to classify 64.8% of the input space) would replicate with the same leverage in a clinical domain where factors are more numerous, factor interactions more complex, and experts potentially less able to render clean binary judgments on abstract scenarios. The empirical validation of the framework in healthcare with new primary data remains an open task.

A second limitation concerns the monotonicity property itself, and here clarification is needed that goes beyond the standard caveat about domain applicability. The efficiency gains reported in Section 3.1 depend not only on monotonicity holding in the abstract but on **the questions having been designed with monotonicity in mind**. A critical function of the LLM in Step 2 of Algorithm 1 is to help construct a factor hierarchy whose questions are naturally interpretable as factors where improving any factor cannot plausibly worsen the decision outcome. The LLM-assisted question generation experiments in Section 4.1 show that this is achievable, and the questions Perplexity AI generated are already structured in ways compatible with monotone reasoning. But this compatibility is a design outcome, not a property that was discovered to hold in a pre-existing question set. An analyst who applied the monotonicity-based elicitation procedure to questions assembled without regard for this property would find the inference rules providing far fewer automatic classifications, and in the worst case generating inconsistencies. The framework's efficiency therefore depends on the quality of the LLM-assisted question design phase in a way that the efficiency figures alone do not reveal. Practitioners deploying the framework should understand that designing monotone-compatible questions is part of the methodology, not a precondition that can be assumed.

In the proposed methodology, experts should be capable of answering generalized questions without requiring detailed follow-up questions. The design process for generalized questions enables them to achieve this level of generalization, as case study 1 demonstrates. However, for some tasks, this level of generalization may not be attainable, which is a limitation of the approach.

Third, the aggregation logic experiments in Section 4.1.5 confirm the limitation of LLM-generated decision rules in a way that the framework addresses but that has not been stated with sufficient directness. The three rules the LLM proposed—majority, weighted, critical threshold—are all drawn from a small family of simple arithmetic aggregators, and none of them is the max formula that FIPS 199 uses and that case study 2 shows is the appropriate rule in the cybersecurity context. The space of plausible aggregation functions over a multi-factor domain is vastly larger than the set of rules LLMs currently propose, as the broader aggregation literature confirms [Dujmovic, 2018; Beliakov et al., 2020]. The monotone elicitation approach handles this by replacing the search for the correct formula with a direct elicitation of the expert's function—but this substitution requires expert time and introduces its own challenges when experts express uncertainty or inconsistency. The framework does not solve the aggregation problem; it reframes it as a structured elicitation problem, which is more tractable but is not cost-free and is not immune to the quality of the expert's introspective access to their own reasoning.

A fourth limitation involves the single-expert assumption that underlies most of the framework's efficiency analysis. The worked example in Section 3.1.1 and the proposal submission case study both implicitly assume a single authoritative expert. In practice, organizational decisions involve multiple stakeholders with partly different mental models. Algorithm 2 addresses group aggregation but in relatively schematic terms, and the combination of multiple EMMs into a group model has not been subjected to the same formal efficiency analysis as the single-expert elicitation. When expert disagreement is substantial—for example, when one expert treats project alignment as a necessary condition whose absence is disqualifying while another treats it as one strong factor among several compensable ones—the discussion and model revision process described in Section 3.2 may require iterations whose cost is difficult to estimate in advance and whose outcome depends on the quality of the facilitation. The framework provides the conceptual tools for surfacing these disagreements (through the 2EMM, explanation model, introduced in Section 1.4) but resolving them remains a human cognitive task that the mathematical framework supports without replacing.

Finally, the comparison in Section 4.1.1 between LLM-generated and human-generated questions found broad comparability but also identified at least one case—question 11 on potential partners—where the LLM's generic formulation missed a crucial dimension that a domain expert would naturally include. In an automated mode without expert review, these gaps would propagate into the EMM and produce a model that is internally consistent but externally incomplete. The framework's insistence on expert review at Steps 2 and 3 of Algorithm 1 is therefore not a procedural formality but a substantive safeguard. The efficiency gains from LLM-assisted factor generation are conditional on this review taking place, and organizations that omit it to save time risk constructing an EMM that is missing the exact factors that matter most in their specific context.

## Implications for Practice and Actual Deployment

The analysis across the three case studies suggests a practical principle for when the investment in EMM construction is justified. The return is highest when the decision is **recurrent** rather than one-time, so that upfront construction cost is amortized across multiple applications; when errors on true positives are costly enough that the conditional misclassification rates documented in Section 4.1.8 are unacceptable; when **expert knowledge is genuinely irreplaceable by available documents or training data**; and when explainability and auditability are requirements rather than desiderata, either for organizational accountability or for regulatory compliance. Most high-stakes organizational decisions in science, security, and clinical care satisfy several of these conditions simultaneously.

The recurrence argument is illustrated most clearly by the proposal submission scenario. A research team that submits proposals regularly faces the same fundamental decision structure repeatedly across different funding opportunities. An EMM constructed during one cycle can be applied to subsequent calls without further expert involvement, requiring revision only when the team's strategic priorities or organizational context change substantially. The mammography case study cited in Section 4.3 demonstrates that elicitation for an 11-dimensional binary factor space can be completed in under an hour; the seven-factor proposal example in Section 3.1.1 suggests similar timescales for domains of comparable dimensionality. Organizations that invest in constructing team-level EMMs would, in effect, be documenting their decision culture in a portable, executable form—one that survives personnel turnover better than informal institutional memory and is more precise than generic checklists or verbal descriptions of organizational priorities.

For cybersecurity authorization, the case for investment is arguably stronger. FIPS 199 already provides a normative framework that many organizations are required to follow, and the ATO process mandates periodic re-evaluation. The EMM framework's contribution is to support the elicitation and documentation of organization-specific impact assessments more systematically, producing a formal, auditable record of the reasoning behind those assessments—one that can be examined by oversight bodies, updated when system conditions change, and compared over time to track drift or deliberate revision. The cost of this construction is modest relative to the documentation overhead of a formal ATO process; the benefit is a living representation of the organization's security reasoning that an informal narrative cannot provide.

For clinical deployment, the evidence reviewed in Section 4.3 demonstrates that expert-extracted monotone functions can outperform purely data-driven models on clinically consequential error types. The framework provides the methodological foundation for validation and refinement—EMMs can be assessed against patient outcomes, compared across clinicians, and updated through additional elicitation cycles—but translating that foundation into a deployable clinical decision support tool involves validation and integration requirements discussed in Section 5.5, and remains a task for future work rather than a deliverable of the present framework.

One dimension of practical deployment that the framework's formal presentation tends to obscure is the qualitative character of the elicitation dialogue itself. The worked example in Section 3.1.1 presents elicitation as a clean sequence of scenarios classified as Y or N. In practice, the dialogue is likely to be more exploratory: experts may express genuine uncertainty about edge cases, request to revisit earlier judgments after encountering a later scenario that reframes their understanding, or discover that their reasoning is more complex and context-sensitive than they initially assumed. The 2EMM, introduced in Section 1.4, is designed precisely for this contingency, providing a mechanism for unpacking the lower-level factors that drove a specific conclusion. Practitioners should plan for the elicitation phase to be a genuine cognitive exercise rather than a mechanical question-answering session and should allocate facilitation support and realistic time accordingly. The one-hour figure from the mammography elicitation represents an achieved result with a cooperative expert and a well-prepared facilitator; it should be understood as a realistic optimum rather than a universal baseline.

## Broader Synthesis

The broader pattern revealed by this paper's theoretical development and case study evidence confirms that the productive integration of cognitive science and LLMs must be conceived as a two-way engagement in which each field constrains and enriches the other, rather than as a unidirectional augmentation of LLMs with human knowledge. The present framework is one concrete instantiation of this bidirectional relationship. LLMs contribute breadth and accessibility—the ability to generate candidate factors, propose structural organizations, and surface considerations that even experienced domain experts might not spontaneously articulate—while the expert mental model contributes depth, specificity, and the decision logic that exists nowhere in any training corpus. Neither is sufficient alone; together, they produce a decision support system that is simultaneously grounded in domain knowledge, responsive to individual expert judgment, computationally tractable, and formally transparent.

The risk of cognitive homogenization identified by Sourati et al. [2026]—the standardization of reasoning through broadly deployed LLMs that reflect and reinforce dominant discourse patterns—is directly countered by the EMM framework's architecture. By grounding decision support in individual expert models rather than population averages, and by preserving the specific reasoning structures of individual practitioners in a formal, inspectable representation, the framework actively resists the homogenization tendency. An organization that maintains its own EMM is, in a meaningful sense, protecting its cognitive distinctiveness against the averaging effect of generic AI systems. The visualization techniques described in Section 3.3, which render the structure of inter-expert disagreement legible rather than suppressing it through aggregation, reinforce this protection: they make visible the cognitive diversity within expert communities rather than dissolving it into a consensus that may not actually exist.

More broadly, the literature synthesized in Section 2 identifies several requirements the field must address: evaluation frameworks capable of assessing causal reasoning and metacognitive monitoring beyond increasingly saturated benchmarks [Center for AI Safety, Scale AI, & HLE Contributors Consortium, 2026; Laurent, 2026a]; domain-specific safeguards targeting the disproportionate hallucination rates documented in expert fields [Lavrinovics et al., 2025]; and scalable tacit knowledge elicitation approaches that maintain the rigor of classical protocols [Ford & Sterman, 1998]. The present framework addresses the last of these directly. The first two motivate the most immediate paper-specific directions for future work: an end-to-end controlled study comparing decision accuracy and hallucination rates between LLM-alone and EMM-guided conditions with real proposals evaluated by real reviewers; formal efficiency bounds for group aggregation analogous to those established for the individual elicitation case in Section 3.1.1; and extension of the elicitation procedure to continuous and context-dependent factors through integration with fuzzy or interval-valued monotone function theory. In each of these directions, the mathematical foundation developed here—efficiency guarantees, formal error bounds, and visualization machinery—provides both the analytical tools and the conceptual vocabulary for progress.

# Conclusion

Important decisions are never made with generic information alone. They are made by real people, in real organizations, at critical moments—drawing on intuition, context, and expertise that often exists nowhere but in their heads. This structural fact is the origin of the problem this paper addresses, and it motivates the paper's central reframing: in the class of decisions where critical knowledge is entirely tacit, LLM hallucination is not a retrieval failure and cannot be cured by better retrieval. It is a failure of model availability. The solution is accordingly not to search harder or align more carefully but to discover and encode the expert's decision model through a principled elicitation dialogue, then provide that model as structured input to the reasoning process. This conceptual move—from augmenting the LLM's access to information to constructing an external, auditable model of expert reasoning—distinguishes the present framework categorically from every prior method for mitigating hallucination.

The three case studies demonstrate a principle of consistent advantage across qualitatively distinct knowledge environments. Where knowledge is entirely tacit, the framework provides the team-specific decision logic that LLMs hallucinate when asked directly. Where knowledge is partially documented, it supplies the organization-specific content that doctrine leaves underspecified. Where prior empirical models exist, it provides the formal structure that makes expert and data-driven models mutually legible, comparable, and correctable. In each environment the LLM is a genuinely useful participant—for generating candidate factors and structuring hierarchies—but it is not sufficient in any of them, and the EMM framework is precisely the complementary mechanism that converts LLM assistance from a source of plausible-but-hallucinated output into a principled component of an accountable decision process.

The framework also directly counters the risk of cognitive homogenization identified in recent theoretical work. By grounding decision support in individual expert models rather than population averages, and by preserving the specific reasoning structures of individual practitioners in a formal, inspectable representation, the approach actively resists the standardizing tendency of broadly deployed LLMs. An organization that maintains its own EMM is, in a meaningful sense, protecting its cognitive distinctiveness against the averaging effect of generic AI systems.

The primary limitation this paper cannot yet resolve is the absence of end-to-end empirical testing in clinical settings and in controlled comparisons between LLM-alone and EMM-guided conditions with real proposals evaluated by real reviewers. The formal analysis of Section 4.1.8 establishes the bounds on hallucination error analytically; instantiating those bounds with observed rather than constructed decision functions remains the most important open task.

The mathematical foundation developed here—efficiency guarantees, formal error bounds, and visualization machinery—provides the analytical tools and conceptual vocabulary for the research directions developed in Section 5.7; the framework this paper establishes does not exhaust that territory, it opens it.